\documentclass{article}

\usepackage[english]{babel}
\usepackage[numbers, sort&compress]{natbib}

\usepackage[letterpaper,top=2cm,bottom=2cm,left=3cm,right=3cm,marginparwidth=1.75cm]{geometry}

\usepackage{amsmath}
\usepackage{graphicx}
\usepackage{rotating}  
\usepackage{subcaption}
\usepackage{booktabs}
\usepackage[table]{xcolor}
\usepackage{tikz}
\usepackage{threeparttable}   
\usepackage{morefloats}
\usepackage{pdflscape}

\definecolor{bestcell}{RGB}{198,239,206}   
\definecolor{worstcell}{RGB}{255,199,206}  
\definecolor{hdr0}{RGB}{220,228,255}       
\definecolor{hdr1}{RGB}{255,235,180}       
\definecolor{hdr2}{RGB}{200,240,210}       
\definecolor{allrow}{RGB}{240,240,240}     
\newcommand{\B}[1]{\cellcolor{bestcell}\textbf{#1}}   
\newcommand{\W}[1]{\cellcolor{worstcell}#1}            
\usepackage[colorlinks=true, allcolors=blue]{hyperref}

\title{The Power of Light: Improving Synthetic-to-Real Domain Adaptation through Physically-Based Indirect Illumination}

\author{
  Hooman Tavakoli Ghinani\thanks{German Research Center for Artificial Intelligence (DFKI), Kaiserslautern, Germany}\ \thanks{RPTU Kaiserslautern-Landau, Kaiserslautern, Germany}
  \and Tatjana Legler\footnotemark[1]\ \footnotemark[2]
  \and Martin Ruskowski\footnotemark[1]\ \footnotemark[2]
}

\begin{document}
\maketitle

\begin{abstract}

While synthetic data generation resolves the manual labeling bottleneck in computer vision, minimizing the syn-to-real domain gap requires optimizing rendering variables. This paper presents a systematic study analyzing the impact of lighting configurations and background complexity on object detection performance. We introduce SmartSDG, an automated, reproducible pipeline built on NVIDIA Isaac Sim using Physically-Based Shading (PBS), alongside ILLUM\_INTRUCK, a new multi-object industrial benchmark dataset. Through 18 controlled experiments utilizing a state-of-the-art YOLOv12 framework, we demonstrate that complex, indirect lighting configurations paired with domain-relevant background variability significantly increase visual cue richness. Our quantitative findings show that avoiding direct specular peaks preserves crucial surface textures, mitigates the domain gap, reduces false positives, and accelerates model convergence compared to using conventional direct-light synthetic data. Ultimately, we provide actionable virtual scene design guidelines to maximize object detection robustness in industrial automation.
\end{abstract}

\section{Introduction}
Object classification and localization are fundamental capabilities of artificial intelligence (AI) and computer vision, with applications spanning diverse domains such as surveillance, autonomous driving, medical imaging, and industrial automation. In industrial contexts, reliable detection and localization of objects provide significant benefits across multiple fields, including human–robot interaction, assembly processes \cite{smallODTavakoli}, worker training, and the advancement of automation in robotics \cite{surati2021pick}. Object Detection (OD) methods aim to both identify and localize target objects within images \cite{rakhimkul2019autonomous, sekkat2021vision}. Among the various approaches, the YOLO (You Only Look Once) framework \cite{redmon2016you} has gained substantial attention due to its real-time inference capability and superior performance compared to other detectors. Since its introduction, YOLO has undergone rapid development, with YOLOv12 now representing one of the state-of-the-art models, offering multiple model sizes optimized for different deployment scenarios \cite{tian2025yolov12}.

Training OD models requires large-scale annotated datasets, where high-quality labels provide the supervision necessary for robust performance in real-world scenarios. However, dataset annotation remains one of the most resource-intensive phases of the pipeline: time-consuming, error-prone, and costly \cite{korakakis2018short}. The challenge lies not only in producing sufficient labels but also in ensuring that the dataset is representative and informative for the detection task at hand \cite{tavakoli2021small}. As shown in prior research, datasets play a central role in the success of deep learning for computer vision \cite{braun1805eurocity}. To achieve high-performance OD, datasets must therefore be diverse, high quality, and as free from bias as possible \cite{jaipuria2020deflating}.

Synthetic data offers significant potential to overcome the bottlenecks associated with real-world datasets. By eliminating the need for manual labeling, synthetic data enables greater control over bias and labeling accuracy, thereby reducing errors, cost, and labor intensity. Virtual environment–based data generation makes it possible to produce photorealistic images together with perfectly aligned annotations, which can then be used to train object detection models. However, despite these advantages, synthetic data introduces its own challenges. The most prominent issue is the domain gap between synthetic imagery generated in virtual environments and real-world scenarios \cite{ghinani2025synthetic, tsirikoglou2020survey}. To mitigate this gap, domain randomization techniques are widely employed during the training process. These techniques involve systematically varying background conditions, illumination, camera positions, object textures, and other characteristics to improve generalization \cite{zhu2025domain}. Among the parameters that can be tuned for domain adaptation, lighting conditions represent a particularly critical factor \cite{minciullo2021db}. 

In virtual environments, adjusting lighting conditions to closely resemble real-world scenarios has been extensively studied. However, most prior work has primarily focused on direct illumination from light sources. To address this limitation, our study introduces a novel approach that emphasizes more realistic lighting configurations, in which light rays undergo multiple interactions with the scene before reaching the objects, thereby more faithfully approximating real-world illumination. We demonstrate that incorporating such complex indirect lighting conditions in synthetic environments leads to improved performance of object detection models.

To evaluate this hypothesis, we propose a dedicated dataset in the industrial domain, specifically targeting manual assembly and human-robot interaction scenarios, such as pick-and-place tasks. The dataset is designed to minimize variations in all factors except lighting conditions and background complexity. We introduce the Illumination Interlock Truck (ILLUM\_INTRUCK) dataset, which captures a variety of lighting conditions as well as virtual environments. Furthermore, we develop a vision-related Synthetic Data Generation (SDG) pipeline, called SmartSDG, using NVIDIA Isaac Sim \cite{NVIDIA_Isaac_Sim}, enabling systematic dataset creation under the controlled industrial live research platform SmartFactoryKL \cite{SFKL24}, a manufacturer-independent, continuously operated testbed for Industry 4.0 applications.

This work presents a comprehensive study on generating optimal synthetic data for object detection, focusing on the light condition, with the goal of reducing the domain gap between synthetic and real data while minimizing energy consumption in terms of both human labor and computational resources. The contributions of this paper can be summarized as follows:

\begin{itemize}

\item We introduce a novel analysis of the effect of multi-bounce, indirect complex lighting configurations versus simple, conventional direct illumination for domain gap mitigation.

\item We investigate how background variations, alongside these novel indirect lighting profiles, improve model robustness and systematically reduce the syn-to-real domain gap.
\item We design and introduce SmartSDG, a synthetic data generation pipeline tailored for systematically studying the complex interplay of indirect multi-bounce lighting physics and background clutter.
\item We generate a synthetic dataset ILLUM\_INTRUCK from the SmartFactoryKL \cite{ZUEHLKE200814101} prototype, explicitly designed as a benchmark to evaluate lighting profiles and background effects on domain adaptation.
\item We provide actionable guidelines on how to design virtual scenes in SDG, specifically focusing on optimizing indirect light transport to maximize OD performance for task-specific industrial applications.
\end{itemize}

\section{Related Work}

Object detection has greatly benefited from advances in synthetic data, where target objects are placed within images in randomized or semi-randomized configurations and paired with precise annotations that accurately specify their positions within the scene \cite{westerski2024synthetic}. Johnson et al. \cite{johnson2016driving} demonstrated that training neural networks on photorealistic computer-generated images from simulation engines can outperform human-labeled datasets. Their experiments, conducted on the KITTI dataset \cite{Geiger2012CVPR} for vehicle detection, highlighted the potential of synthetic data to reduce reliance on costly manual annotation. 
In the context of object detection, domain randomization has emerged as a well-established approach to addressing the bottleneck of limited data availability. Relating to the assumption of a finite, even large, distribution of possible object appearances, such as placements, rotations, and other variations, the core idea is to systematically randomize scene parameters in order to generate datasets that better approximate this distribution and, in turn, improve model generalization. However, in practice, a domain gap often arises due to the difficulty of accurately mimicking this distribution \cite{westerski2024synthetic}.

Tremblay et al. \cite{Tremblay_2018_CVPR_Workshops} explored different domain randomization techniques, including variations in lighting, object pose, and textures, and proposed a pipeline for training object detection models on synthetic data. Their findings showed that networks pretrained on synthetic data and fine-tuned on real datasets achieved superior performance compared to models trained solely on real data. Using cars as target objects, their evaluation was performed on the KITTI dataset, with random lighting and viewpoint variation explicitly incorporated into the SDG process.

More recently, Tavakoli et al. \cite{ghinani2025synthetic} investigated the combination of synthetic data with active learning for fine-tuning across multiple benchmark and industrial datasets. The study demonstrated that this hybrid approach effectively mitigates the domain gap between synthetic and real data while significantly improving object detection performance. Dwibedi et al. \cite{dwibedi2017cut} proposed an approach in which object instances are extracted as patches from real images and blended into randomized backgrounds. They argued that the realism of the object patch itself is more important than the realism of the entire image, and demonstrated that combining synthetic data generated from these patches with a small portion of real data can yield better performance compared to models trained solely on real data. Their assumption was that such blending enables neural networks to focus on object-specific details rather than boundary artifacts, which often arise from lighting variations or imperfections in the image-cutting process.
In a survey on synthetic data for object detection, Westerski et al. \cite{westerski2024synthetic} categorized the variables used in domain randomization into several groups, including foreground, background, distractor objects/occlusion, camera parameters, lighting, noise, and blur. Their analysis revealed that most research efforts primarily focus on foreground generation when applying domain randomization.

Hinterstoisser et al. in \cite{hinterstoisser2018pre} show that simply freezing the feature extractor layers of a pretrained model trained on real images and further fine-tuning on synthetic data can enhance the performance on the R-CNN architectures. In their work, they used clutter background images for SDG, and a small perturbation of the light color was part of the domain randomization in their study. They showed that the performance was not considerably improved by the random light color.
Tremblay et al. in \cite{Tremblay_2018_CVPR_Workshops}  demonstrated a domain randomization approach in which a random number of lights of different types were inserted at random locations, with scenes rendered from random camera viewpoints and subsequently composited over random background images. Their ablation study highlighted the critical role of lighting variation: when lights were randomized but brightness and contrast augmentations were disabled (“no light augmentation”), the average precision (AP) dropped slightly. In contrast, when training was performed under a fixed lighting setup (“fixed light”), the AP decreased significantly, underscoring the importance of randomizing lighting conditions.
Nogues et al. \cite{nogues2018object} present domain randomization in SDG using variations in different parameters, including textures, camera positions, and the number of lights and their colors. Hinterstoisser et al. \cite{hinterstoisser2019annotation} introduced a synthetic dataset generation pipeline that leverages 3D models as backgrounds, incorporating domain randomization across factors such as light position and color. Their experiments demonstrated that blurring and the light color randomisation had the most pronounced impact on model performance.
Wrenninge et al. \cite{wrenninge1810synscapes} introduced the Synscapes street-scene synthetic dataset, offering photorealistic images for training models with lighting behavior simulated through unbiased path tracing \cite{kajiya1986rendering}. The dataset employs physical sun and sky models in combination with physically based reflectance models to capture realistic surface interactions, thereby producing more natural and diverse lighting conditions.
Schraml et al. \cite{s24020649} emphasized lighting conditions as a pivotal factor in synthetic data generation and systematically evaluated their impact on AI performance. Their results demonstrated that incorporating diverse lighting conditions significantly improved classification accuracy, showing the importance of lighting diversity for enhancing model generalization.

Eversberg et al. \cite{eversberg2021generating} also mentioned the importance of lighting in SDG. In their study, the authors compared different domain randomization strategies by evaluating AP on industrial blade object detection tasks using synthetic images with varying levels of realism and variability. In particular, they examined two lighting models, point lights and image-based lighting, and reported that image-based lighting with HDRIs achieved slightly higher AP compared to point lights.
Borkman et al. \cite{borkman2021unity} proposed a Unity-based package for synthetic data generation, called the Perception package. They also evaluate the object detection for synthetic data generated employing directional lights to illuminate scenes with randomized intensity and color. Mao et al. \cite{mao2021domain} incorporated lighting variations within their domain randomization framework to encourage the detection model to focus on fine-grained features of the target object (bird class).

Given the importance of light emission in synthetic dataset generation using virtual environments, we refer to the formulation of light behavior in the rendering process as described by Kajiya \cite{kajiya1986rendering}:

\begin{equation}
L_o(x, \omega_o) = L_e(x, \omega_o) + \int_{\Omega} f_r(x, \omega_i, \omega_o)\, L_i(x, \omega_i)\, (n \cdot \omega_i)\, d\omega_i
\label{equition}
\end{equation}

\begin{flalign*}
L_o(x, \omega_o) &: \quad \text{Outgoing radiance} && \\[2mm]
L_e(x, \omega_o) &: \quad \text{Emitted radiance} && \\[1mm]
f_r(x, \omega_i, \omega_o) &: \quad \text{Bidirectional Reflectance Distribution Function (BRDF)} && \\[1mm]
L_i(x, \omega_i) &: \quad \text{Incoming radiance} && \\[1mm]
(n \cdot \omega_i) &: \quad \text{Geometric term} && \\[1mm]
\Omega &: \quad \text{Integration hemisphere} &&
\end{flalign*}

In this study, we distinguish between direct and indirect illumination based on the light transport described by the rendering equation. Direct illumination refers to radiance arriving at a surface directly from an explicit light source without intermediate interactions. In the proposed SmartSDG pipeline, direct illumination is provided by multiple rectangular ceiling-mounted area lights positioned within the virtual factory environment. Indirect illumination, in contrast, arises when incident radiance undergoes one or more inter-reflections between scene surfaces before reaching the target object. These interactions are computed through physically based global illumination, where light is repeatedly reflected by the surrounding environment, including walls, floors, machinery, and other factory components. The presence of this complex scene geometry enables multiple light transport paths, producing realistic indirect illumination, soft shadow transitions, and subtle color bleeding that improve the visual fidelity of the generated synthetic images.

The image formation process is described using the rendering equation~\eqref{equition}, which models the transport of light in a scene. In this formulation, illumination arises from the interaction of emitted and reflected radiance across scene surfaces. The incoming radiance contributes to surface shading through direct and indirect components computed by a global illumination solver that accounts for multiple light bounces within the environment. In the proposed synthetic data generation pipeline, all objects are non-emissive; therefore, the emitted radiance term $L_e(x, \omega_o)$ is set to zero.

\section{Our Study}

In our study, we employed YOLOv12 as the object detection pipeline to evaluate the proposed methodology. 

\subsection{Experimental Setup}
A series of experiments were designed, each consisting of two camera configurations for capturing images of the component(s) within the field of view. The first camera (Camera 1) was positioned in a ceiling-like setup, capturing images from a semi-vertical perspective, coaxial to the vector of the overhead light source, which is called coaxial illumination in this paper. The second camera (Camera 2) was positioned at the nominal vertical height of the target component(s) to capture direct frontal profiles. Because this camera's perspective is perpendicular to the primary overhead illumination, we refer to this setup as off-axis illumination throughout this paper. To systematically investigate the role of lighting in domain adaptation and domain gap mitigation, we defined three controlled light-intensity settings: Low (L), Medium (M), and High (H), while maintaining a fixed light-source position and color. The intensity values, color temperature, and spatial positioning of the overhead sources were empirically captured based on conventional SDG parameters for SmartFactoryKL to accurately resemble the real-world illumination on the testbed.

\begin{itemize}
    \item \textbf{Experiment 0:} In this baseline setup, only the target component(s) are present in the scene, with no background visible from either camera point of view. Camera~1 captures a semi-top-down view of the component directly illuminated by ceiling lights, while Camera~2 captures the component from a side view.
    
    \item \textbf{Experiment 1:} In this experiment, a ground plane from the SmartFactoryKL environment is introduced. Camera~1 captures images with a factory floor as the background that includes light reflections, whereas Camera~2 captures side-view images. 

    \item \textbf{Experiment 2:} This experiment is designed to place both cameras, lights, and target component(s) within the full virtual environment of the SmartFactoryKL. Camera~1 captures with coaxial illumination with the factory floor as background, while Camera~2 captures with off-axis illumination with the SmartFactoryKL environment as the background.  
\end{itemize}

\subsubsection{Theoretical Alignment with the Rendering Equation}
The design of these experiments is grounded in the rendering equation (Eq. \ref{equition}). Experiments 0 and 1 focus on the simplest form of light transport, where emitted radiance ($L_e$) is zero, and incoming light ($L_i$) is primarily direct. In contrast, Experiment 2 captures the full complexity of the equation by maximizing the integral term ($\int$) over $\Omega$. By introducing sophisticated environmental geometry, the framework enriches the indirect illumination profiles via multi-bounce diffuse reflections. This scattered ambient radiance illuminates the target objects, thereby elevating the spatial and contextual fidelity of the synthesized visual cues.

\subsubsection{Isolation of the variation}
To isolate the effects of lighting conditions from background complexity, we strategically positioned two cameras to create comparative baselines. In Experiment 0 and Experiment 1, Camera 2 maintains a consistent black, empty background, allowing for a direct assessment of how light interacts with the objects themselves. Furthermore, Camera 1 in Experiment 0 is directly comparable to Camera 2 in Experiment 1; this setup enables a study of coaxial versus off-axis illumination while keeping the background constant. Similarly, for Camera 1, we utilized the same white plane background across Experiment 1 and Experiment 2 to evaluate varying light complexity under a stable environment. Finally, in Experiment 2, Camera 2 introduces both a variable background and complex lighting, a combination that ultimately leads to the superior performance observed in our results.

\subsubsection{training and testing setup}

In our study, YOLOv12 was trained and evaluated using an NVIDIA A100 GPU. The training pipeline was configured with the following empirical and conventional hyperparameters: initialization from pretrained \texttt{yolov12l} weights (pretrained on the COCO \cite{lin2014microsoft} dataset) to ensure a high-capacity model, an input resolution of 640$\times$640 pixels following standard vision benchmarks, and a batch size of 32 to guarantee stable gradient optimization. All other hyperparameters were left unchanged and used the default configurations provided by Ultralytics \cite{tian2025yolov12}. Training was sustained for 200 epochs with an early-stopping patience of 100 epochs; this conservative threshold is conventionally applied in SmartFactoryKL benchmarks to guarantee complete model convergence and prevent premature termination during late-stage optimization. 

Crucially, all default data augmentation techniques were disabled to isolate the evaluation entirely to the synthetic lighting dynamics and background clutter variations.

\subsection{SmartSDG }
In this study, we introduce \textit{SmartSDG}, a synthetic data generation package specifically designed to support the experiments conducted. The package facilitates the creation of distinct datasets that systematically capture variations in domain factors such as lighting conditions, scene clutter, and complex backgrounds. The pipeline is highly extendable and can be adapted for the SDG in the SmartFactoryKL operational demonstrator. Moreover, it provides automated conversion of annotations into YOLO-compatible labels and includes a filtering mechanism that excludes images in which target objects are occluded by more than $50\%$. The entire process is executed in headless mode, allowing sample generation to be reproduced without requiring GUI access. This pipeline eliminates the overhead of manual scene manipulation, such as adjusting distances, locations, and orientations of objects. The overview of the SmartSDG configurations is provided in the table \ref{tab:smartsdg}.
The SmartSDG package is built upon NVIDIA Isaac Sim. Aside from the specific experimental configurations described in this study (Table \ref{tab:smartsdg}), all other environmental and rendering parameters remain consistent with the default NVIDIA Isaac Sim settings. By utilizing the engine's native physically-based shading (PBS) techniques, which approximate the physical interaction between light and materials, and its native path-tracing capabilities without external modification, our results are reproducible and directly attributable to the controlled variables of lighting and scene complexity.

\subsection{Dataset}
\begin{figure}[ht]
    \centering
    \begin{tabular}{cccc}
        \includegraphics[width=0.22\textwidth]{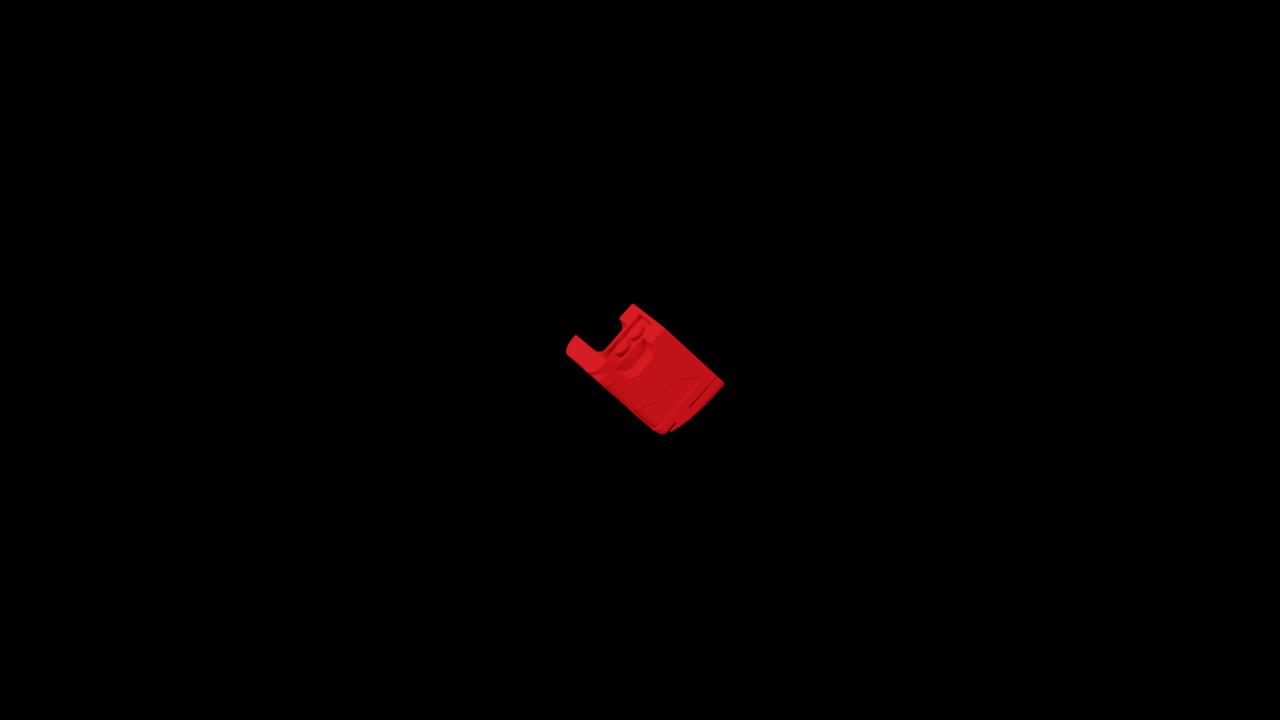} &
        \includegraphics[width=0.22\textwidth]{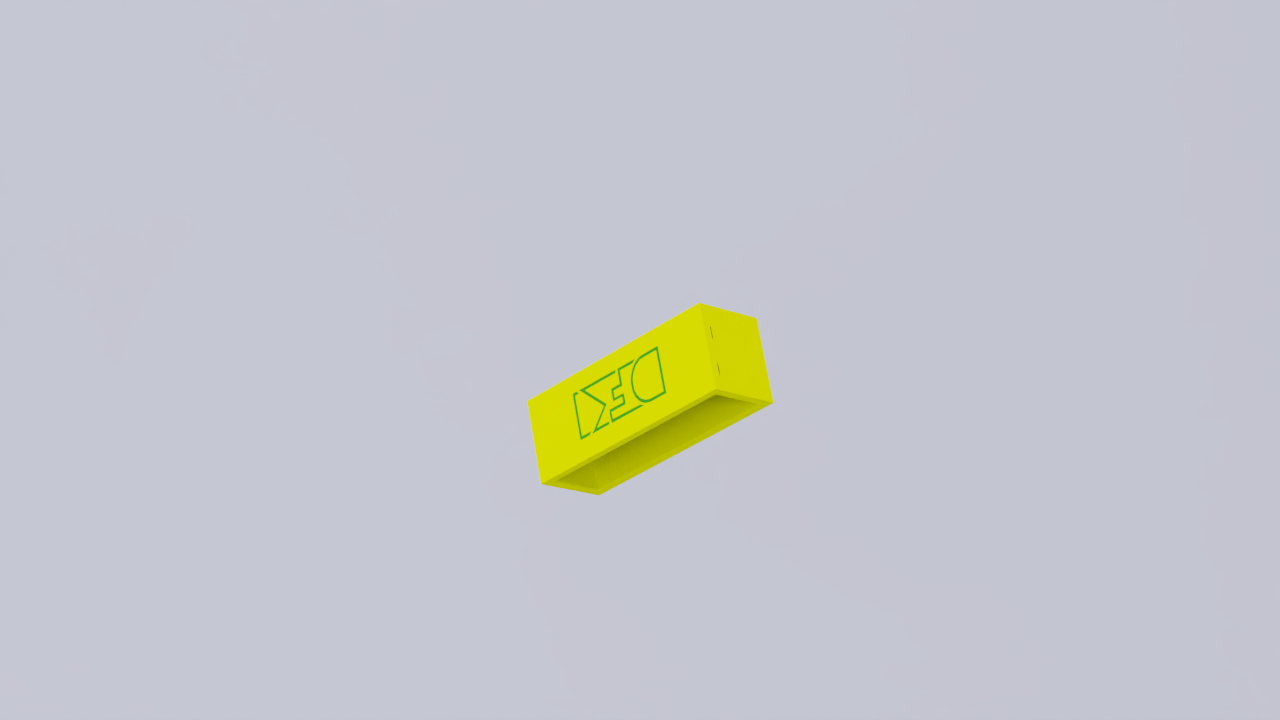} &
        \includegraphics[width=0.22\textwidth]{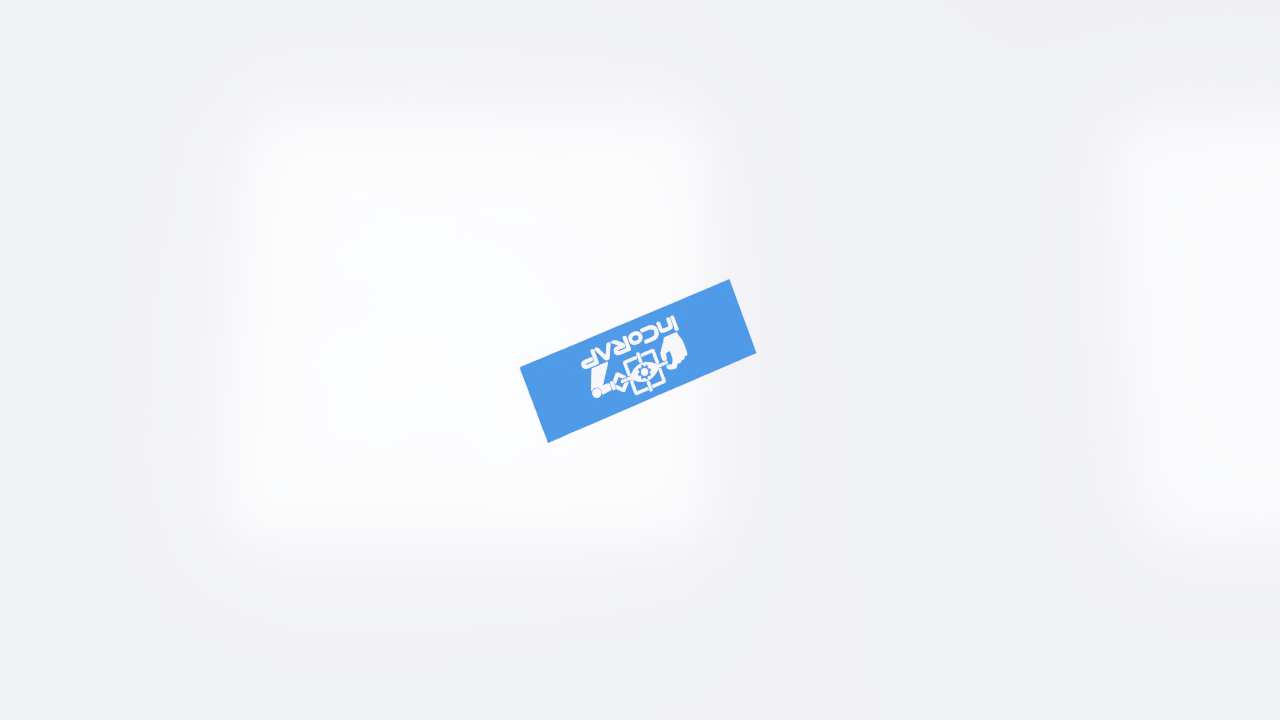} &
        \includegraphics[width=0.22\textwidth]{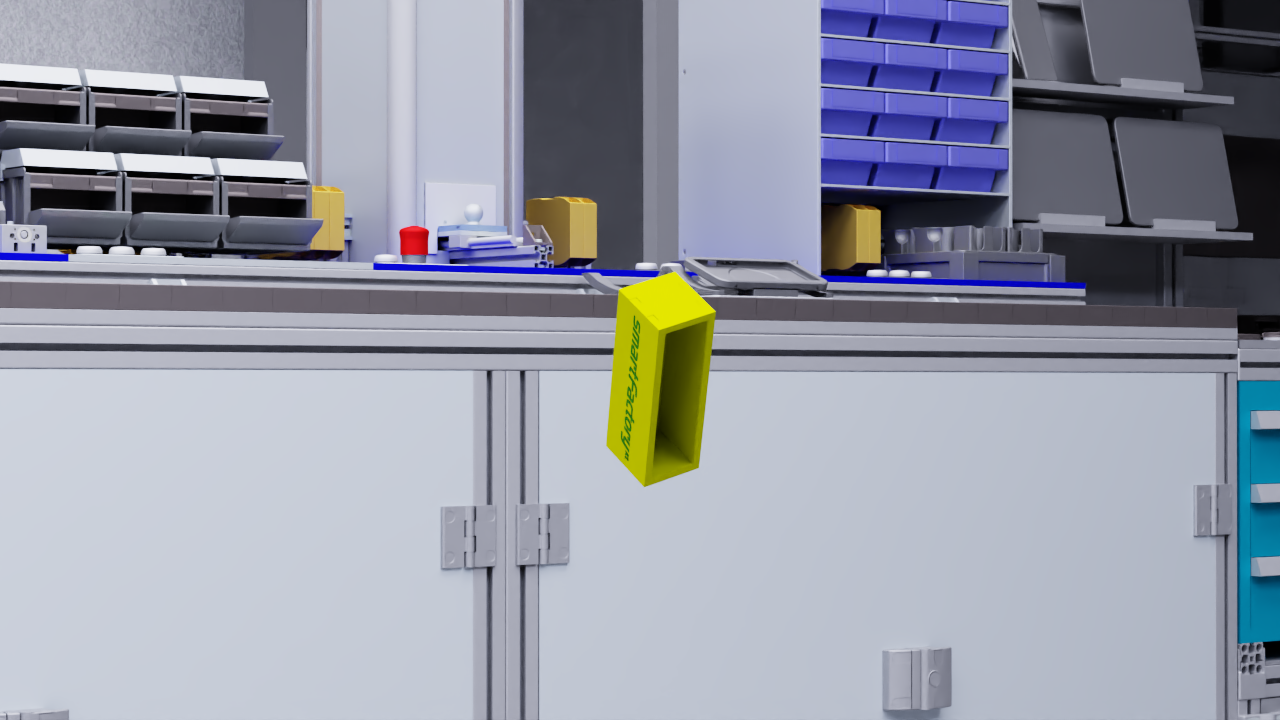} \\

        \includegraphics[width=0.22\textwidth]{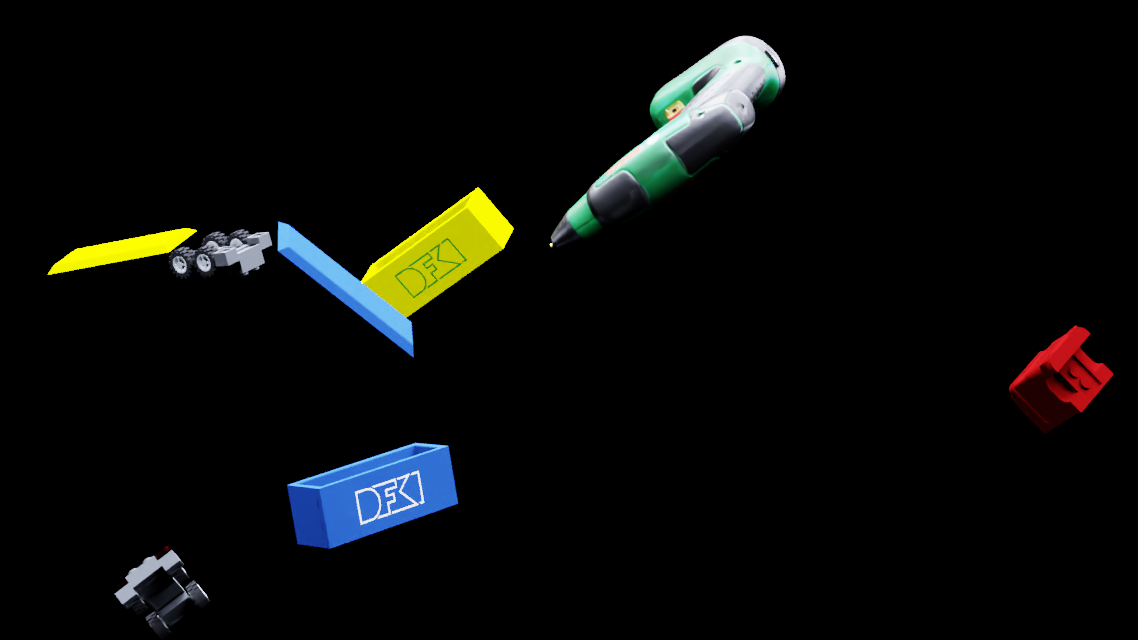} &
        \includegraphics[width=0.22\textwidth]{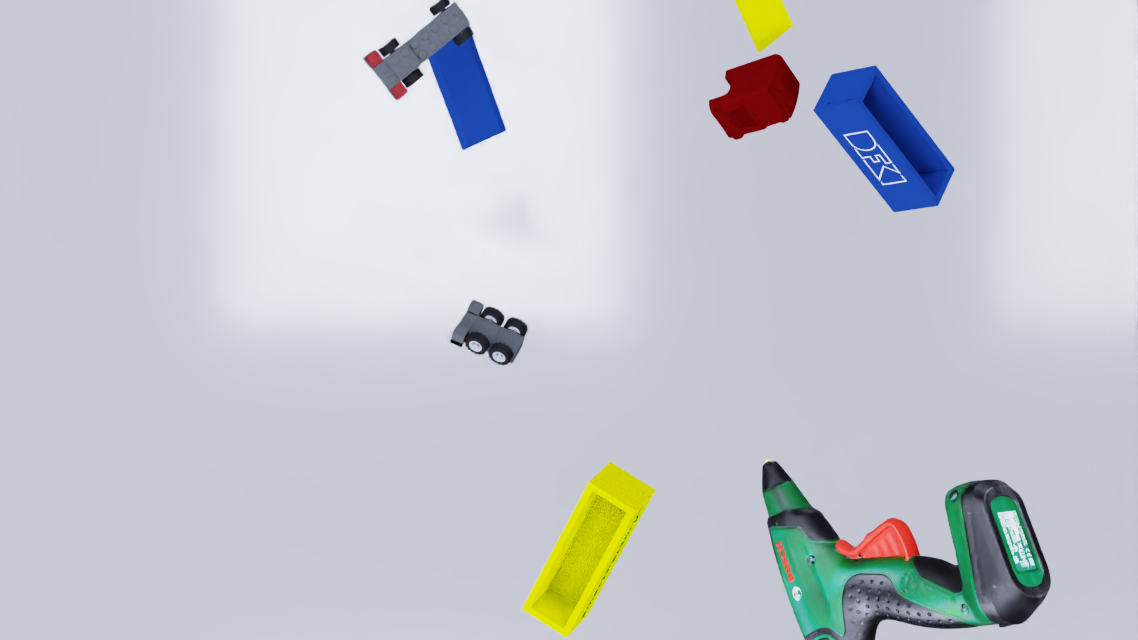} &
        \includegraphics[width=0.22\textwidth]{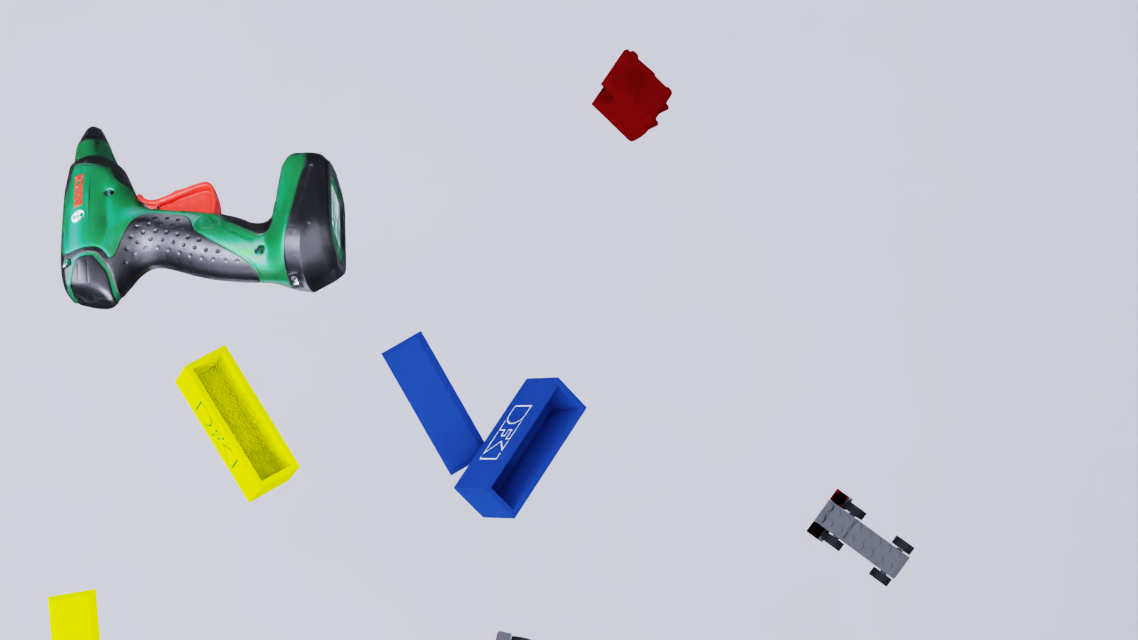} &
        \includegraphics[width=0.22\textwidth]{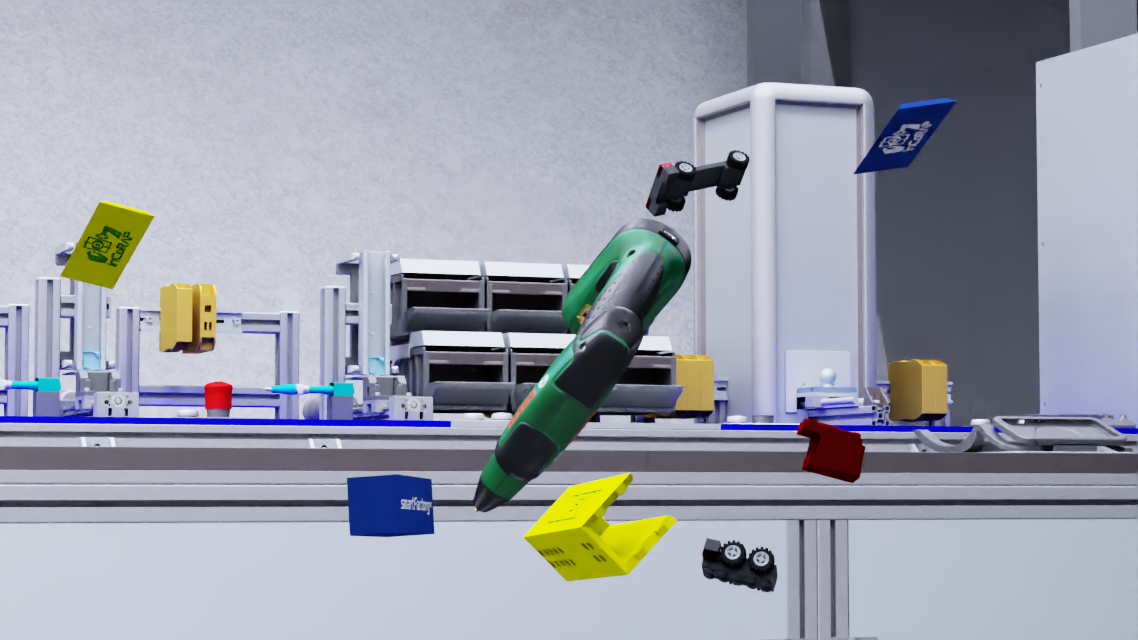} \\
        
        \small (a) Black Background & \small (b) Simple White Plane & \small (c) Simple White Plane & \small (d) Complex Background \\
    \end{tabular}
    \caption{\textbf{Sample images from the ILLUM\_INTRUCK dataset showing different components under varying scene configurations (camera viewpoints, lighting conditions, and backgrounds).} The first row shows single-component images, while the second row depicts multi-object images from different experiments.}
    \label{fig:dataset_samples}
\end{figure}

The dataset consists of 200 images per component, captured at constant distances comparable to the inference scenario of the object detector. With 8 component classes, this results in a total of 1,600 single-object samples. To ensure maximum visual detail, these 200 images per class were generated using high-quality ray-casting settings, prioritizing the preservation of fine textures and complex light-surface interactions (Table \ref{tab:smartsdg}).
In addition, 3,000 multi-object samples were generated, ensuring a balanced distribution across classes and reducing bias. While these 3,000 images were also generated using ray casting, the rendering parameters (such as samples per pixel and max bounces) were adjusted to manage computational time without sacrificing the core visual cues required for training.
Crucially, these rendering settings were kept identical across all experiments and datasets to ensure that performance variations are attributable only to changes in lighting and background, rather than differences in image quality.
Altogether, each experimental setup (combination of camera view, lighting condition, and scene configuration) contains approximately 4,600 images. The samples in Figure~\ref{fig:dataset_samples} illustrate a subset of the dataset generated with SmartSDG, showcasing different components under varying scene configurations.

The computational cost of generating images in Isaac Sim relies heavily on hyperparameter configurations. Consequently, our SDG framework employs distinct parameters for single- and multi-object scenes (Table \ref{tab:smartsdg}). Given the same hardware, high-fidelity single-object generation takes approximately 2.5 minutes per image, whereas multi-object generation takes only around 15 seconds. To balance dataset size with image quality, we utilize lower-fidelity multi-object generation to rapidly scale the number of annotated samples. Meanwhile, high-fidelity single-object images are retained to help the OD model capture essential texture and background features.

\begin{figure}[ht]
    \centering
    \includegraphics[width=\linewidth]{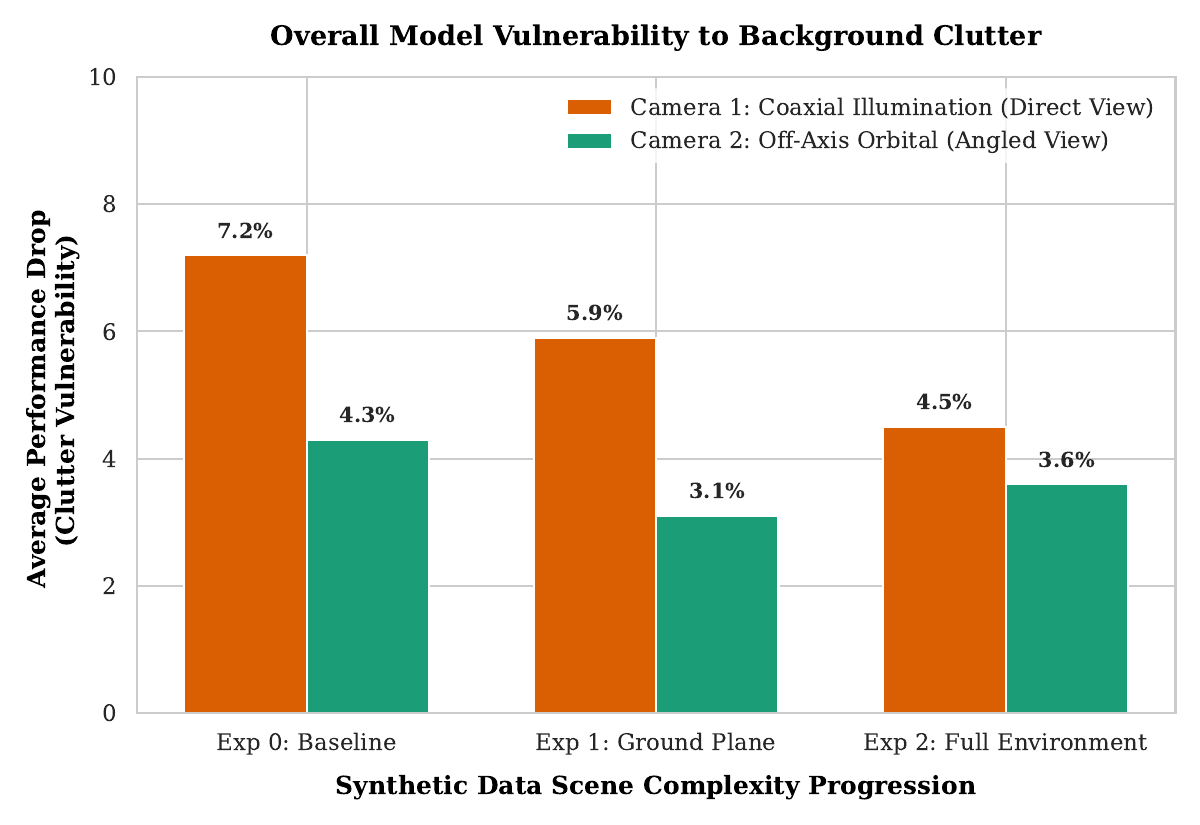}
\caption{\textbf{Aggregated analysis of model vulnerability to background clutter and false positives across experimental phases.} The bar chart illustrates the average performance drop (mAP@[.50:.95] degradation) when evaluating the object detection models on the expanded real dataset (which includes background-only clutter images) relative to the domain-specific baseline. A closer value to zero indicates low vulnerability to clutter and high robustness, meaning the model's detection performance is more reliable under complex conditions. The consistently lower degradation observed for Camera 2 across all stages mathematically demonstrates that the off-axis orbital view (non-coaxial illumination) yields significantly greater stability in real-world environments compared to the coaxial illumination setup.}
\label{fig:merg_real}
\end{figure}

The test dataset consists of 167 real-world labeled images and 140 images that contain only background features with no target objects present. This specific composition is designed to evaluate the model's robustness against False Positives (FP) by challenging its ability to correctly ignore background regions and cluttered industrial environments. By including a high ratio of background-only samples, we can assess the model's discriminative power beyond simple detection accuracy. Additionally, the validation set comprises 20 real-world images, each augmented with three 90-degree rotations, leading to an 80-image validation dataset.

\section{Results}
In this section, we present the results of our study and provide a detailed analysis of the performance of the object detection model trained under each experimental setup. The performance of the object detection models is evaluated using mAP, with results validated on the real dataset.

The overall results are summarized in Table~\ref{tab:results_illustration_improved}, covering all 18 experiments conducted under different lighting conditions, camera positions, and scene configurations. The table reports both the class-wise performance and the aggregated mAP for all classes.
Figure \ref{fig:merg_real} illustrates the aggregated analysis of model vulnerability to background clutter and false positives across experiments.

\begin{figure}[ht]
    \centering
    \begin{subfigure}[t]{0.49\textwidth}
        \centering
        \includegraphics[width=\textwidth]{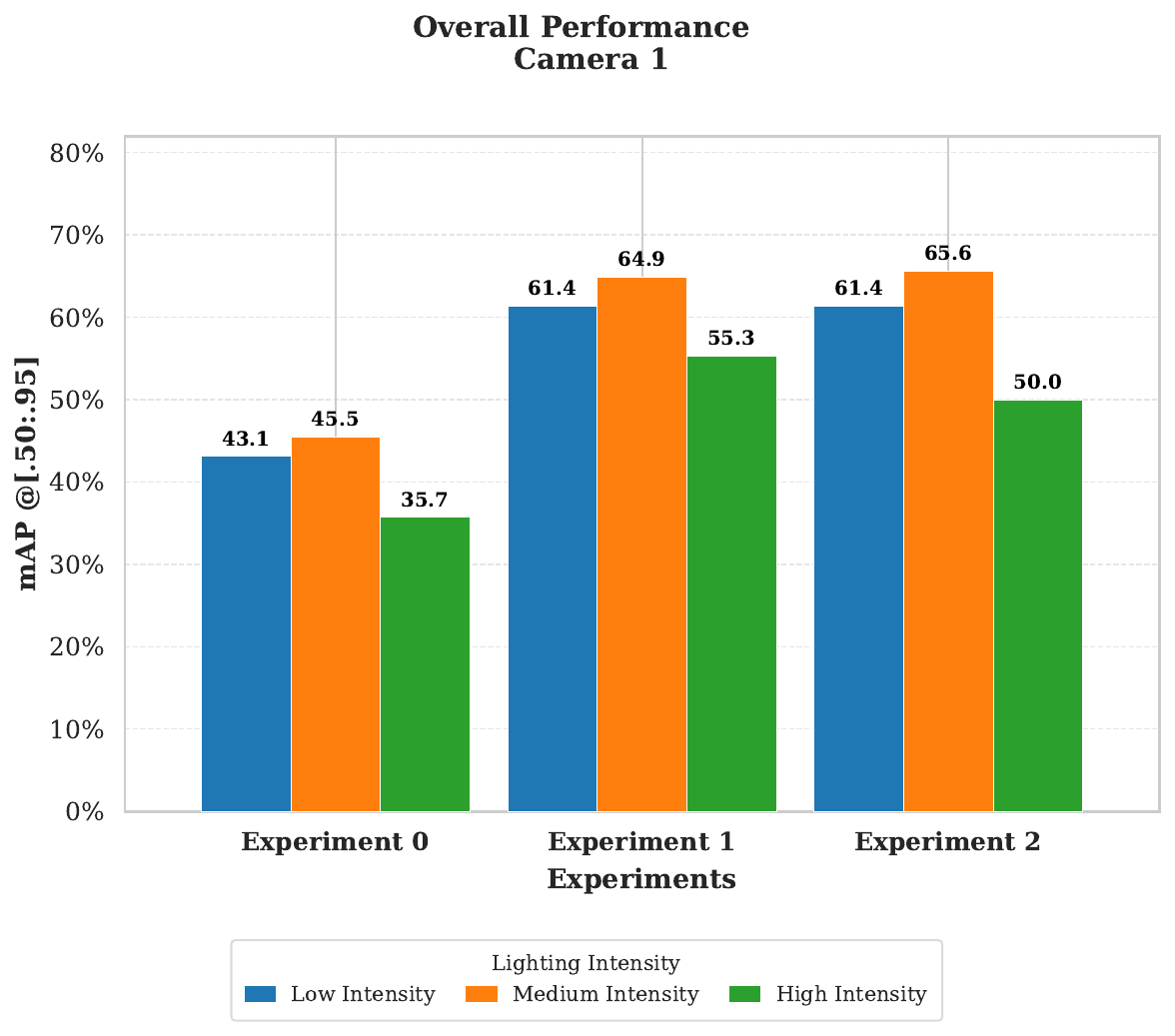}
        \caption{Overall performance on mAP for Camera 1 in a top view across three experiments under varying light intensities.}
        \label{fig:both_diagramsA}
    \end{subfigure}
    \hfill
    \begin{subfigure}[t]{0.49\textwidth}
        \centering
        \includegraphics[width=\textwidth]{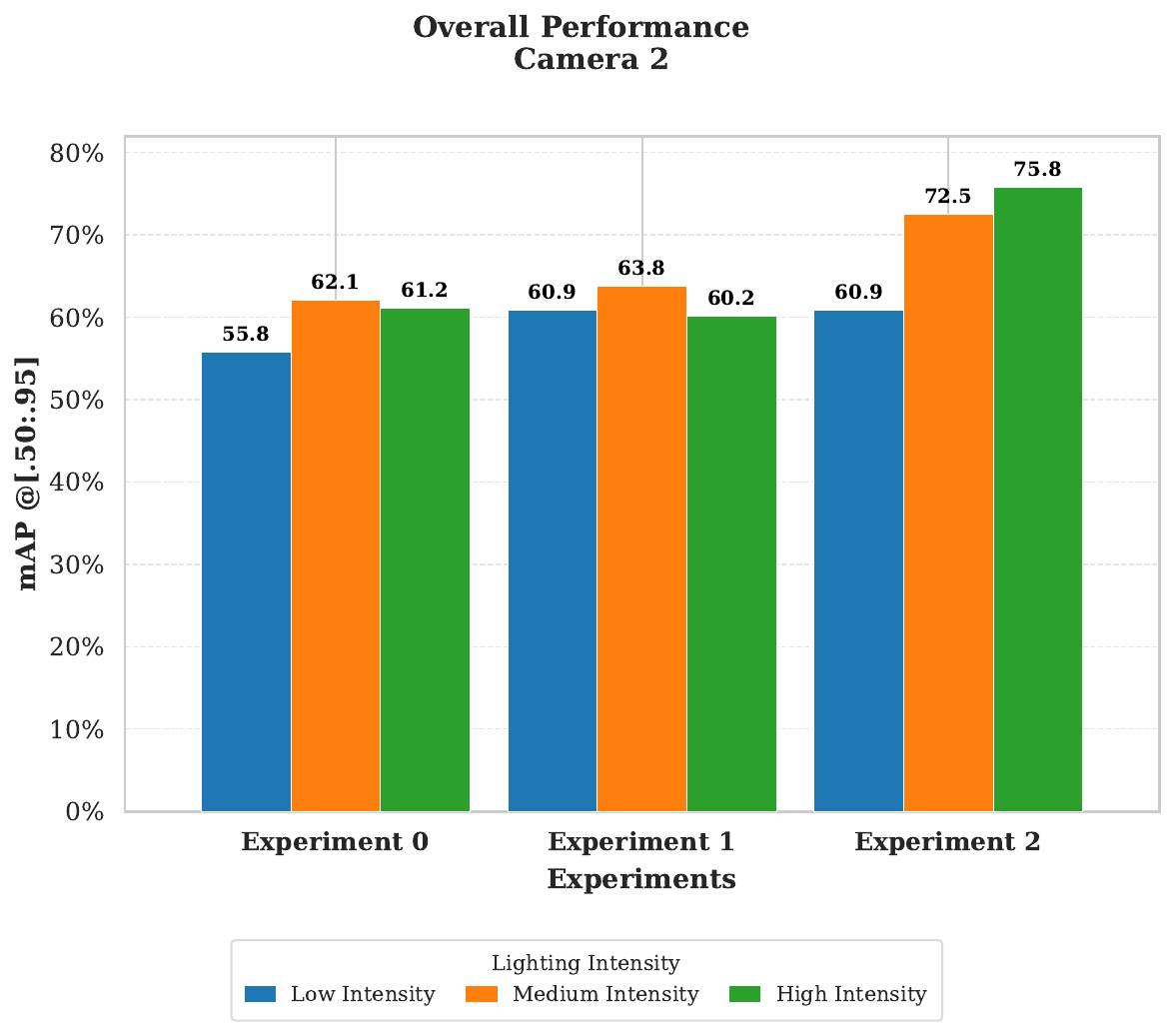}
        \caption{Overall performance on mAP for Camera 2 in an orbital view across three experiments under varying light intensities.}
        \label{fig:both_diagramsB}
    \end{subfigure}
    \caption{An evaluation of the experiments over all classes, with results presented separately for each camera position.}
    \label{fig:both_diagrams}
\end{figure}

\begin{figure}[]
    \centering

    \begin{subfigure}[t]{0.49\textwidth}
        \centering
        \includegraphics[width=\textwidth]{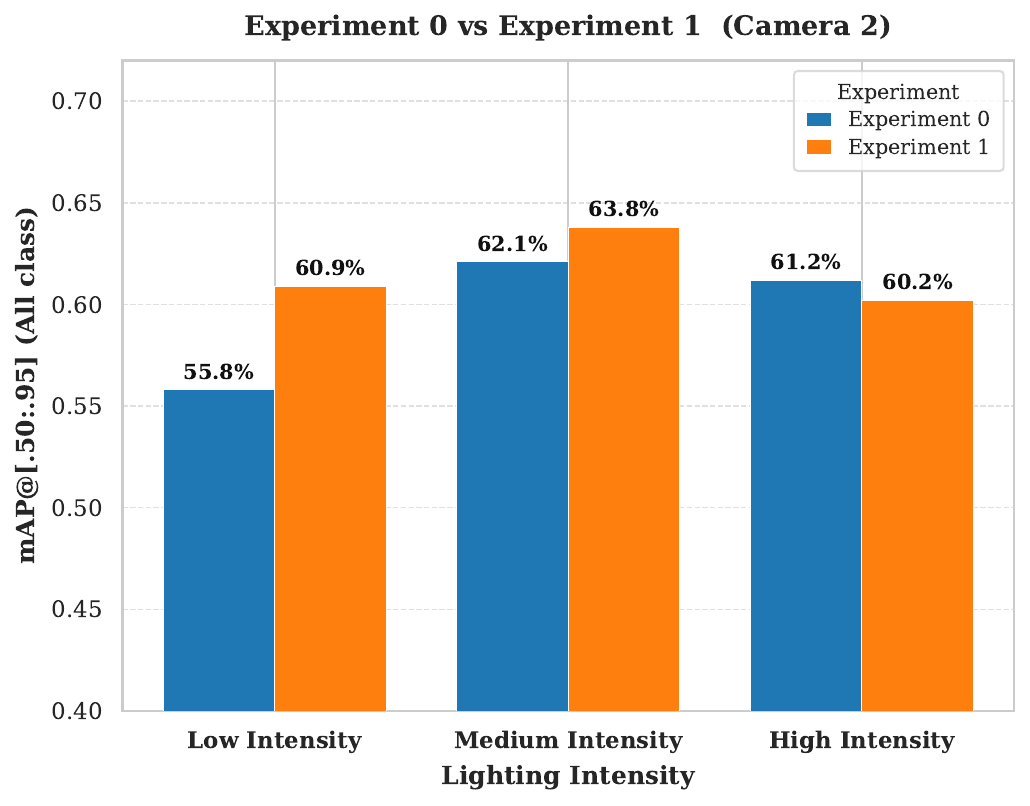}
        \caption{Comparison of Experiment 0 vs Experiment 1 on the Camera 2 dataset (empty background).}
        \label{fig:both_camA}
    \end{subfigure}
    \hfill
    \begin{subfigure}[t]{0.49\textwidth}
    \centering
    \includegraphics[width=\textwidth]{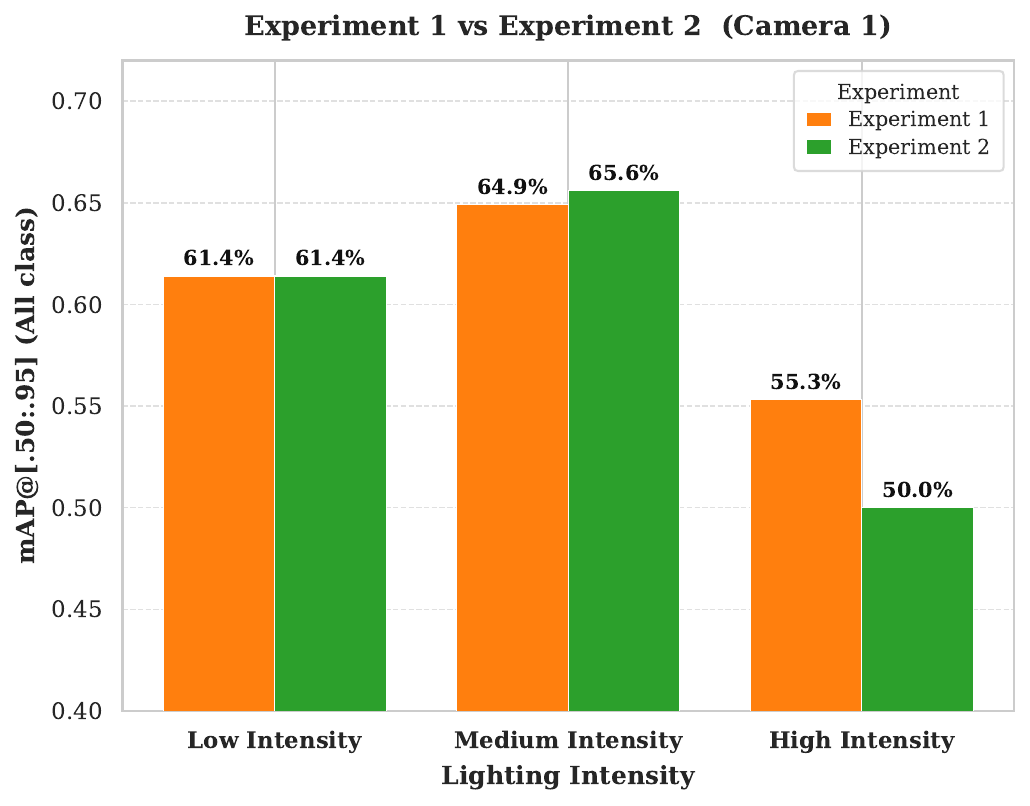}
    \caption{Comparison of Experiment 1 vs Experiment 2 on the Camera 1 dataset (white plane floor background).}
    \label{fig:both_camB}
\end{subfigure}
    \caption{\textbf{Bar chart showing the mAP scores of experiments under various lighting conditions (L, M, H).} The subplot \ref{fig:both_camA} compares Experiment 0 and Experiment 1 on the dataset with an empty background (Camera 2), while the subplot \ref{fig:both_camB} compares Experiment 1 and Experiment 2 on a dataset with a white plane floor background (Camera 1). }
    \label{fig:camera_comparison}
\end{figure}

\section{Discussion}
This section provides an overview of the key findings from our experimental study. We will evaluate and interpret the results derived from our various experimental setups.

A comprehensive comparison of the experimental results is provided in Figure~\ref{fig:both_diagrams}. Subfigure~\ref{fig:both_diagramsA} displays the results for Camera~1, while Subfigure~\ref{fig:both_diagramsB} presents the results for Camera~2.

In the direct vertical view captured by Camera~1 with ceiling lighting, Medium illumination consistently achieves the highest mAP across all three experiments, while Low intensity produces comparable but slightly lower results. A significant and consistent performance collapse is observed at High intensity across all experiments, with Experiment~0 High reaching only $35.7\%$, the lowest result in the entire study. This confirms that direct, high-intensity illumination saturates the synthetic image, causing a loss of fine detail and surface texture. From a computer vision perspective, high-intensity direct lighting produces clipped pixel values where the radiance $L_o$ exceeds the sensor's dynamic range, yielding a near-zero spatial gradient ($\nabla I \approx 0$) across the object surface. Since deep learning models rely on these gradients to extract features such as edges, corners, and surface textures, the suppression of this information prevents the model from identifying discriminative features during training. Furthermore, with coaxial illumination — the camera parallel to the ceiling light source — the specular peak of the BRDF dominates the captured signal, effectively masking the diffuse reflectance components that carry the object's unique visual signatures. Consequently, the model fails to converge on a robust object representation, producing the observed mAP decrease at High intensity.

Sub-figure~\ref{fig:both_diagramsB} reports Camera~2 performance. By positioning its optical axis off-axis relative to the ceiling lights, Camera~2 avoids the specular peak of the BRDF and instead samples the diffuse lobe, preserving natural shadowing and surface textures and producing synthetic training samples that more closely approximate real-world appearance. Camera~2 exceeds Camera~1 in 6 of 9 conditions overall. Camera~1 records marginal advantages in only three conditions, at Low intensity in Experiments~1 and~2 ($+0.5$~ Percentage Point (pp)) and at Medium intensity in Experiment~1 ($+1.1$~pp). These exceptions are confined to indirect lighting configurations, where reduced specular saturation allows both cameras to sample the diffuse lobe with comparable efficiency. Moreover, in Experiment~1, Camera~1 benefits from a smooth white plane, as opposed to the empty black background used for Camera~2. Additionally, Camera~1 performs slightly better under low-intensity lighting in Experiment~2, as the combination of low light and a cluttered background otherwise degrades the results.

Under direct lighting (Experiment~0), Camera~2 outperforms Camera~1 across all three intensity levels by margins ranging from +12.7~pp to +25.5~pp, as Camera~1's coaxial geometry places it within the specular peak of the BRDF, an effect whose severity scales with illumination intensity and becomes catastrophic at high intensity. The advantage is largest in the complex indirect environment: at High intensity in Experiment~2, Camera~2 achieves 75.8\% against Camera~1's 50.0\% (a gain of +25.8~pp), representing the largest camera-geometry effect observed in this study. Comparing Experiment~1 and Experiment~2 through Camera~2 isolates the combined contribution of the complex SmartFactoryKL background and the multi-bounce indirect lighting emitted by the factory environment onto the target objects; Experiment~1 provides a single ground-plane reflector, while Experiment~2 introduces a fully inhabited virtual factory whose surfaces create additional indirect bounces that enrich the diffuse radiance reaching each component. At Medium and High intensities, this compound environmental contribution produces gains of $+8.7$~pp and $+15.6$~pp, respectively, over Experiment~1 Camera~2, confirming that scene-level training diversity and multi-bounce illumination jointly improve domain transfer when sufficient illumination is available to render background features with discriminative contrast.

At Low intensity, however, Experiment~1 and Experiment~2 achieve identical mAP scores for both cameras simultaneously. This establishes a clear minimum illumination threshold: below this level, the physically based renderer cannot resolve the factory environment with sufficient fidelity to effectively train the OD model. Consequently, the complex background degenerates into low-contrast clutter that contributes no additional learning signal compared to an empty background. Furthermore, the low light intensity fails to generate sufficient indirect light bounces from the complex environment for the orbital off-axis camera. This intensity-dependent behavior directly informs the design guideline that complex scene environments should only be deployed with at least Medium indirect illumination intensity.

An analysis of the combined performance trends reveals a compelling crossover effect between sensor geometry, environmental complexity, and illumination intensity. For Camera~1 (coaxial geometry), (Sub-figure~\ref{fig:both_diagramsA}), the system exhibits a strict performance ceiling, peaking uniformly under Medium illumination across all three experiments, a specular trap in which low-intensity profiles generate insufficient feature contrast while high-intensity profiles introduce catastrophic specular glare. Conversely, Camera~2 (off-axis geometry), (Sub-figure~\ref{fig:both_diagramsB}) demonstrates a highly dynamic adaptation to scene physics: in Experiment~1, all three intensity levels perform within a narrow $3.6$~pp range ($60.2$--$63.8\%$), reflecting the limited photometric variation available from an empty background with ground plane as single reflector; upon transitioning to the full industrial environment in Experiment~2, the optimal setting shifts decisively to High intensity, achieving the global maximum of $75.8\%$ as multi-bounce indirect radiance breaks up harsh specular paths and fills object surfaces with rich feature gradients. Ultimately, this study demonstrates that optimising synthetic training datasets requires co-designing camera geometry alongside illumination profiles to match the light-transport properties of the target industrial environment.

\begin{figure}[ht]
    \centering
    \includegraphics[width=\linewidth]{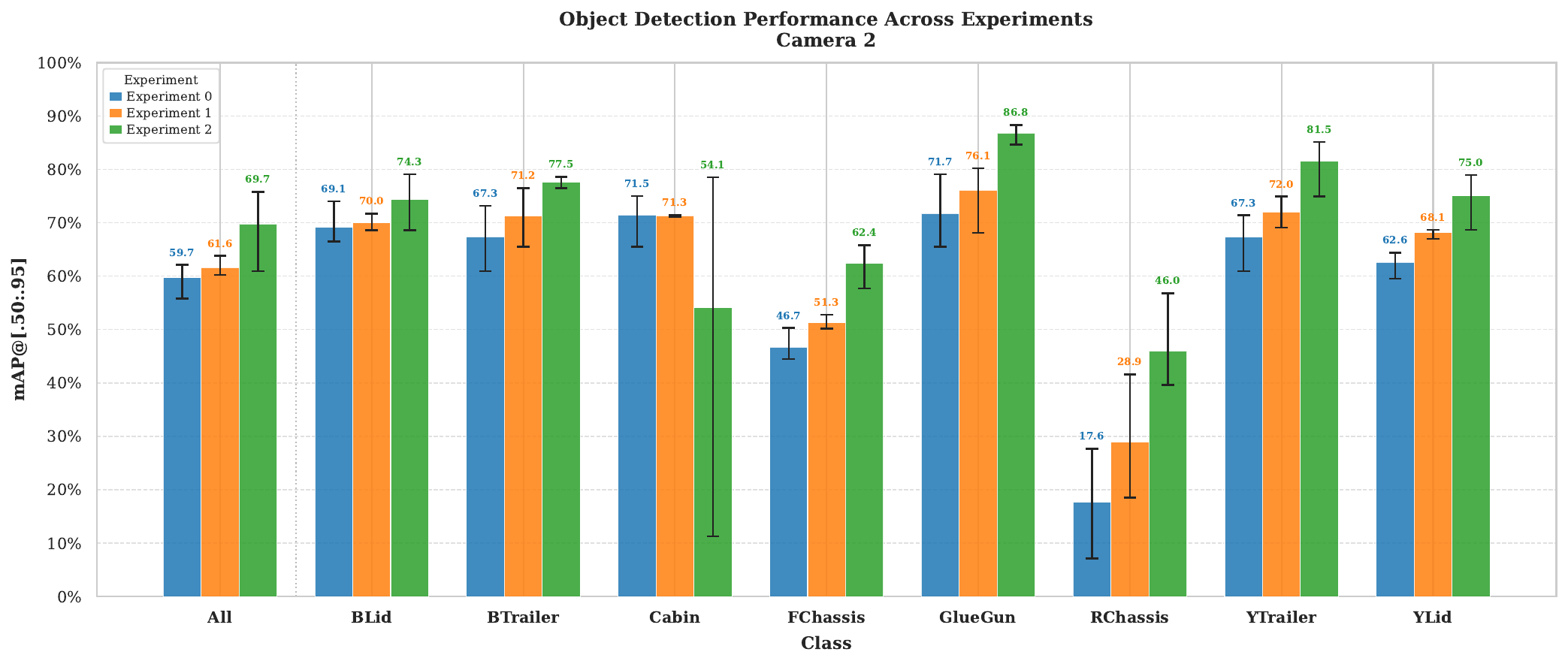}
\caption{\textbf{Class-wise comparison of experiments for Camera~2 on averaging over different lighting intensity levels (High, Medium, Low)}. Whiskers indicate the min--max range across intensities, reflecting the sensitivity of each class to lighting intensity within that experiment. The Cabin class shows a notably wide downward whisker for Experiment~2, corresponding to the Low-intensity collapse ($11.3\%$). RChassis records the widest overall range in Experiment~0, with the minimum reaching $7.1\%$ under Low intensity and a mean of $17.6\%$ mAP.}
\label{fig:avrage_chart_classwise_cam2}
\end{figure}

\begin{figure}[h!]
    \centering
    \begin{subfigure}[t]{0.48\textwidth}
        \centering
        \includegraphics[width=\textwidth]{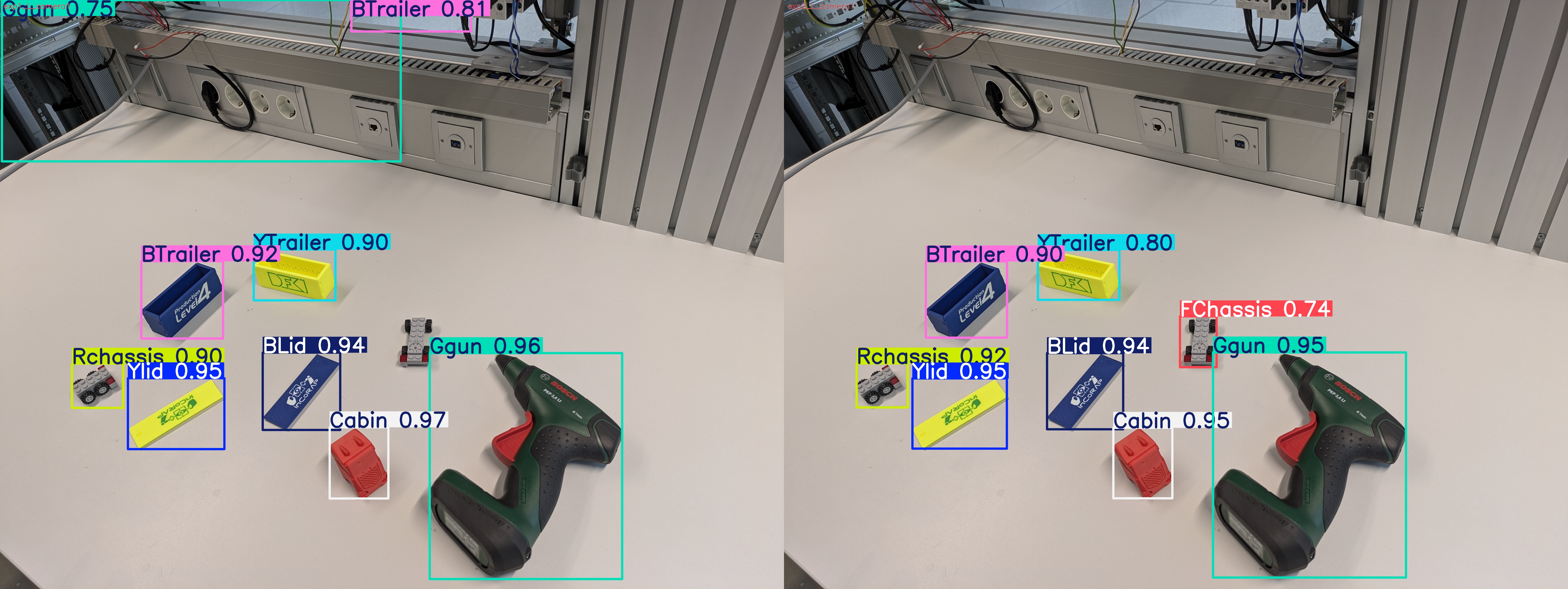}
        \caption{Comparison of experiment 1 vs. 2 with Camera 1 (High intensity)}
        \label{fig:cam1-exp1-2-L}
    \end{subfigure}
    \hfill
    \begin{subfigure}[t]{0.48\textwidth}
        \centering
        \includegraphics[width=\textwidth]{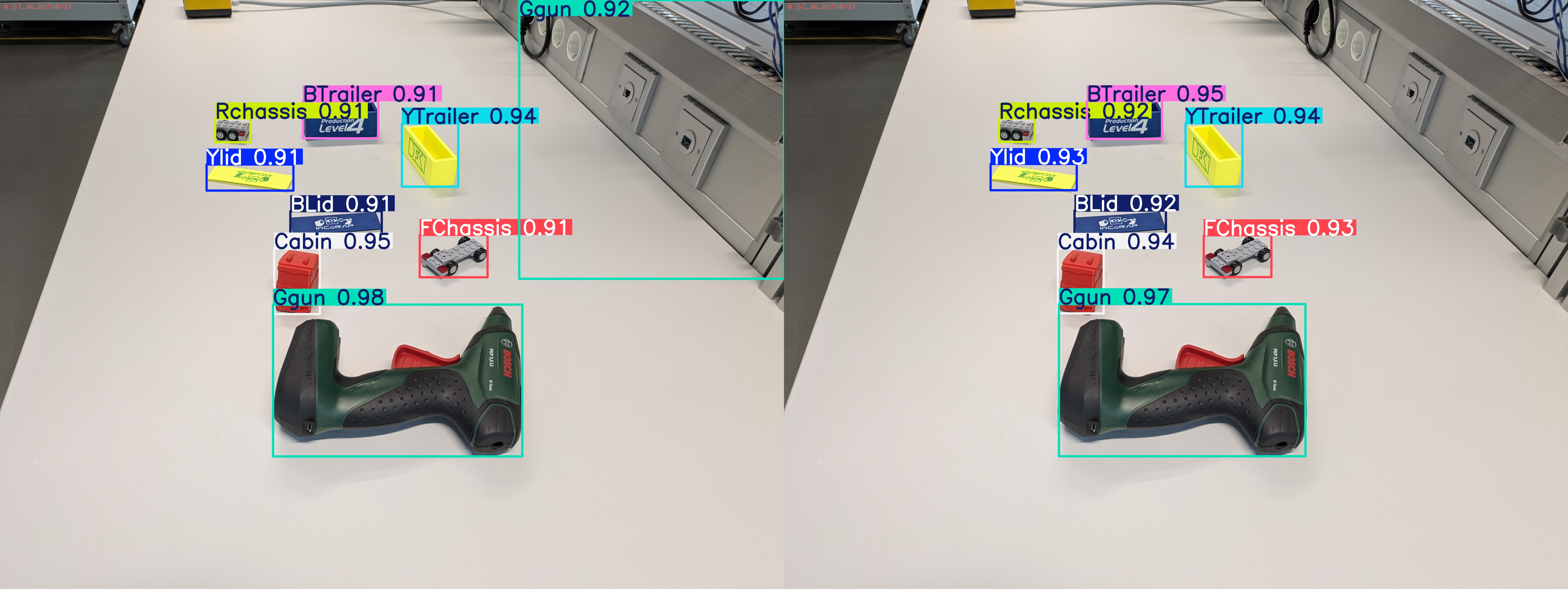}
        \caption{Comparison of experiment 1 vs. 2 with Camera 1 (Medium Intensity)}
        \label{fig:cam1-exp1-2-M}
    \end{subfigure}
    
    \vspace{1em}
    
    \begin{subfigure}[t]{0.48\textwidth}
        \centering
        \includegraphics[width=\textwidth]{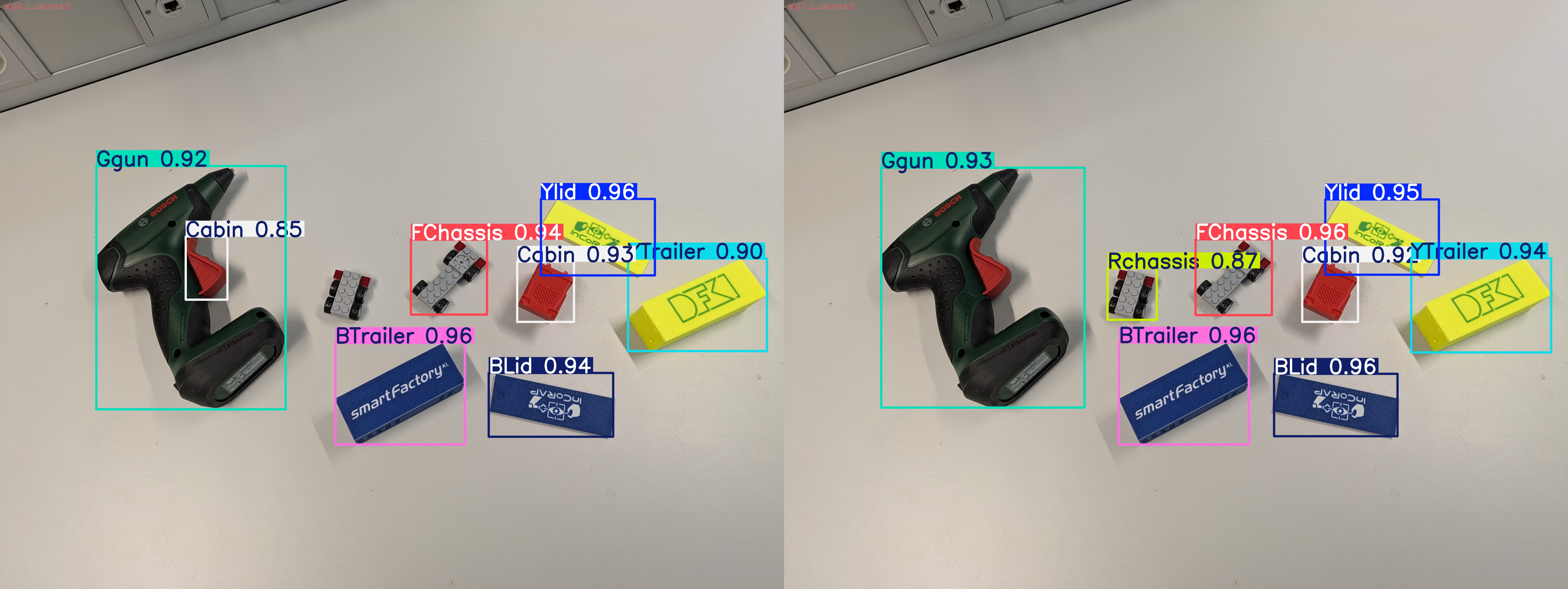}
        \caption{Comparison of experiment 0 vs. 1 with Camera 2 (High intensity)}
        \label{fig:cam2-exp0-1-L}
    \end{subfigure}
    \hfill
    \begin{subfigure}[t]{0.48\textwidth}
        \centering
        \includegraphics[width=\textwidth]{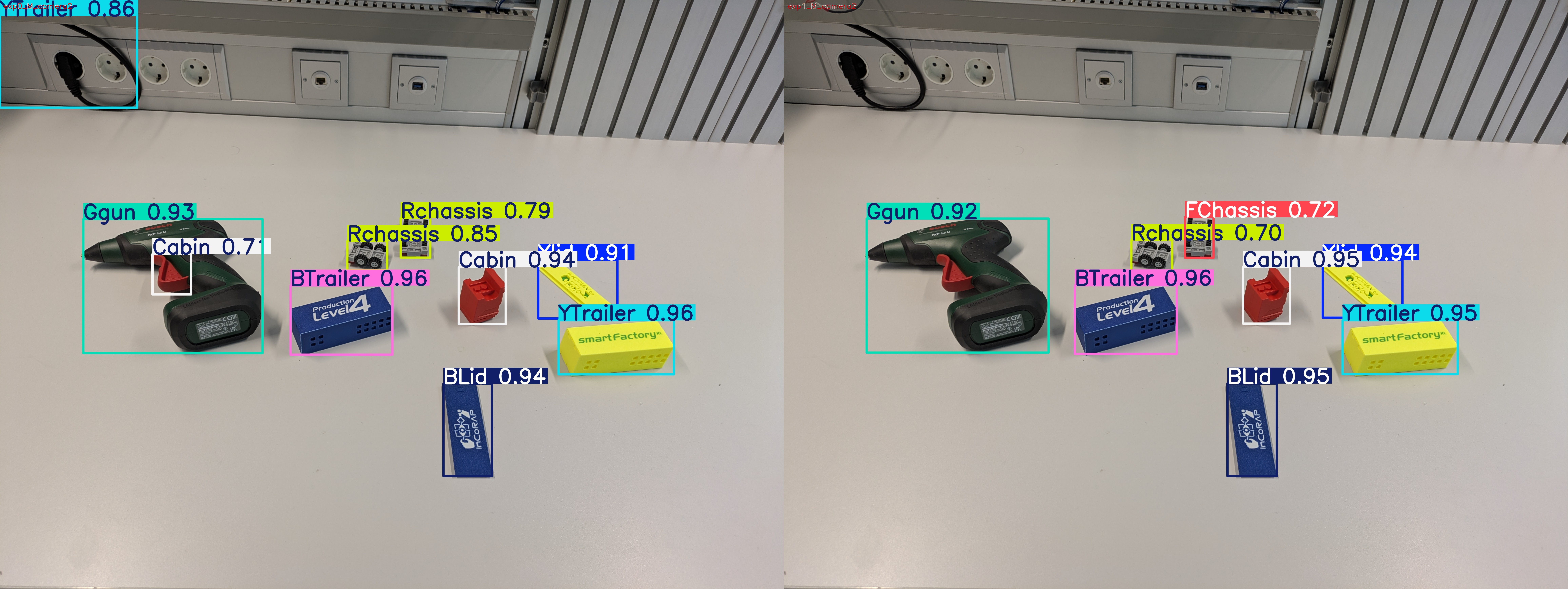}
        \caption{Comparison of experiment 0 vs. 1 with Camera 2 (Medium Intensity)}
        \label{fig:cam2-exp0-1-M}
    \end{subfigure}

    \vspace{1em}

    \begin{subfigure}[t]{0.48\textwidth}
        \centering
        \includegraphics[width=\textwidth]{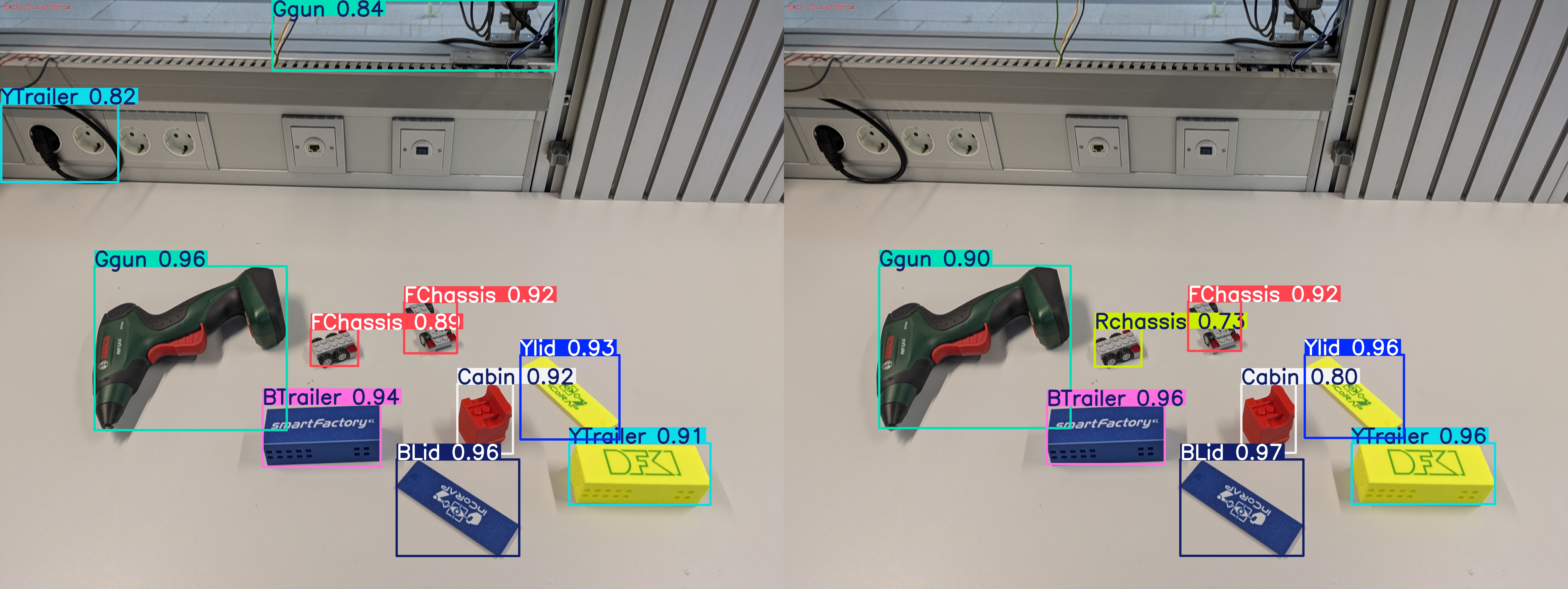}
        \caption{Comparison of experiment 0 vs. 1 with Camera 2 (Low Intensity)}
        \label{fig:cam2-exp0-1-S}
    \end{subfigure}
    \hfill
    \begin{subfigure}[t]{0.48\textwidth}
        \centering
        \includegraphics[width=\textwidth]{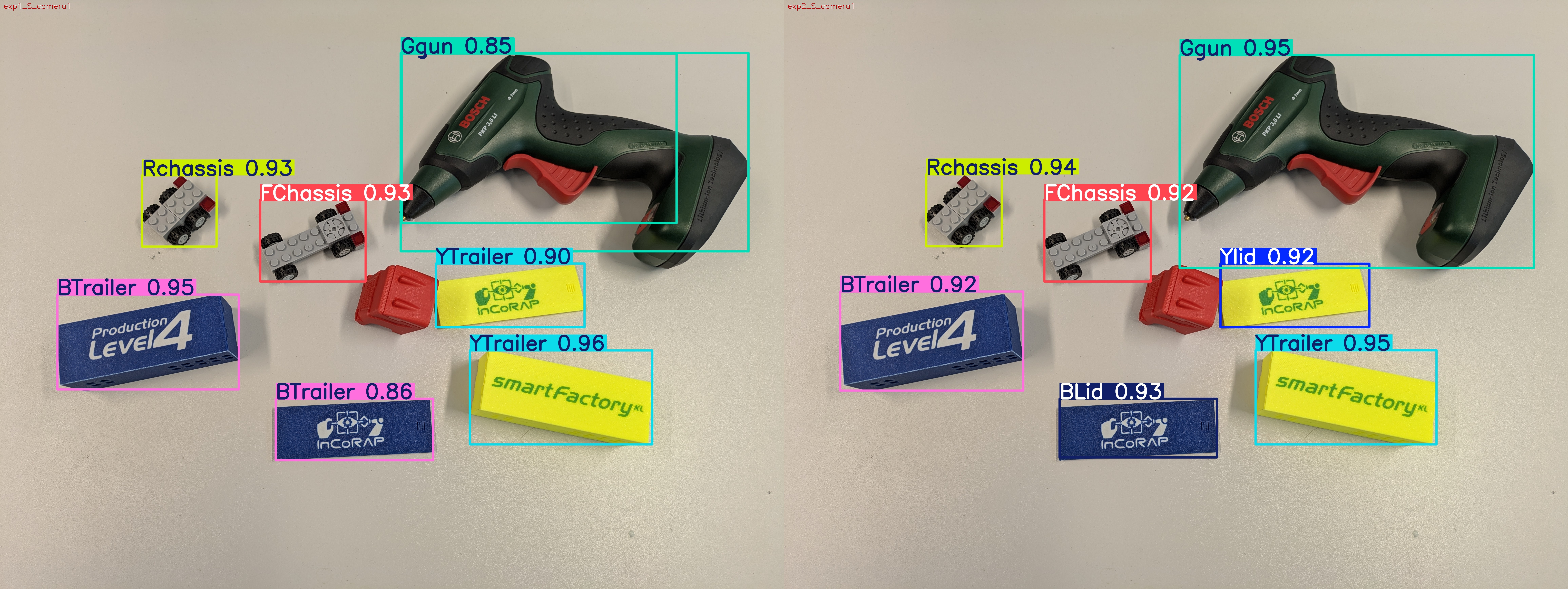}
        \caption{Comparison of experiment 1 vs. 2 with Camera 1 (Low Intensity)}
        \label{fig:cam1-exp1-2-S}
    \end{subfigure}

    \caption{Side-by-side comparison of different experimental setups and light intensities.}
    \label{fig:overall-comparison_images_sidebyside}
\end{figure}

\begin{figure}[h!]
    \centering
    \begin{subfigure}[t]{0.31\textwidth}
        \centering
        \includegraphics[width=\textwidth]{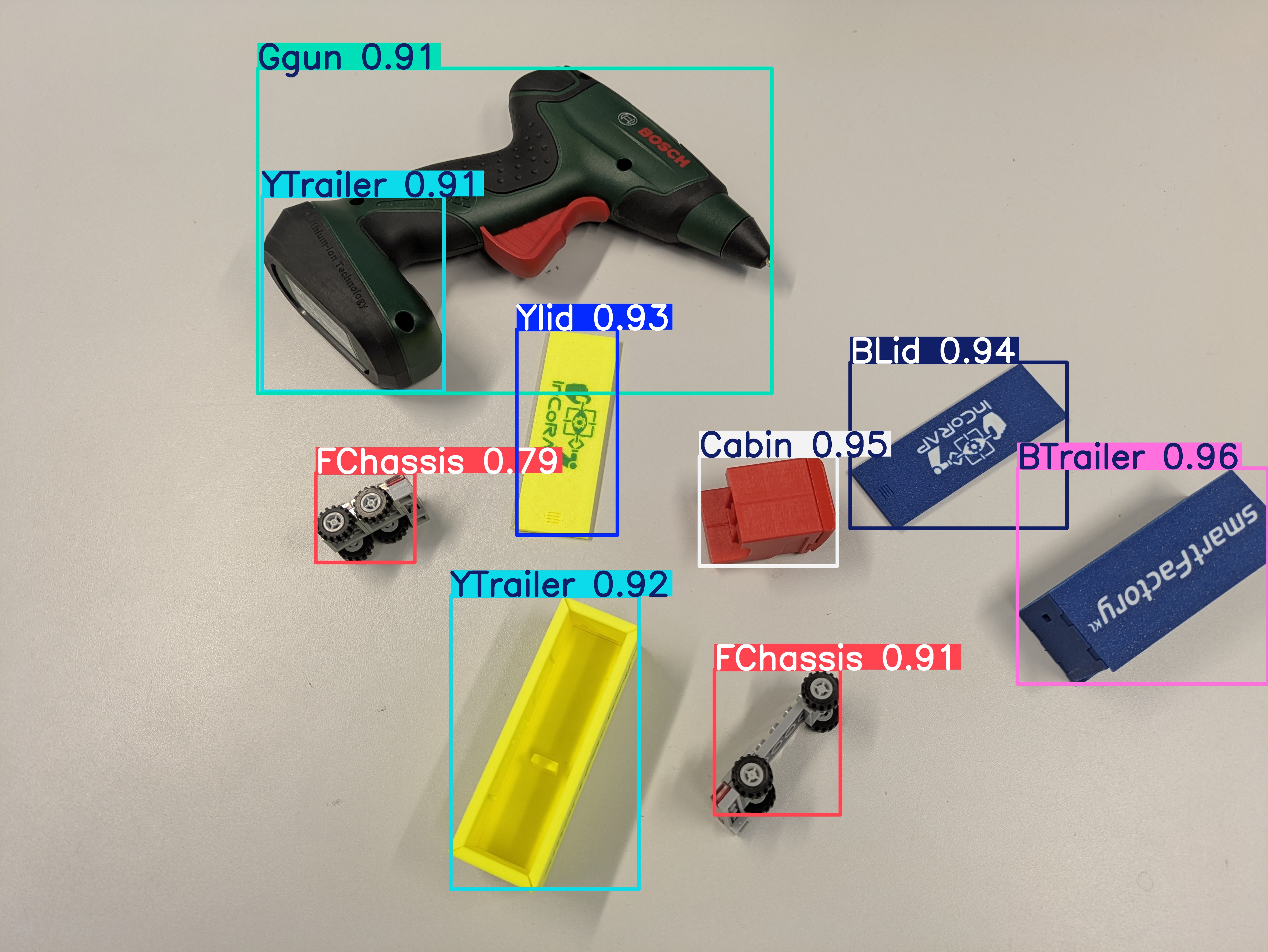}
        \caption{Experiment 0 results}
        \label{fig:exp0-cam2-M}
    \end{subfigure}
    \hfill
    \begin{subfigure}[t]{0.31\textwidth}
        \centering
        \includegraphics[width=\textwidth]{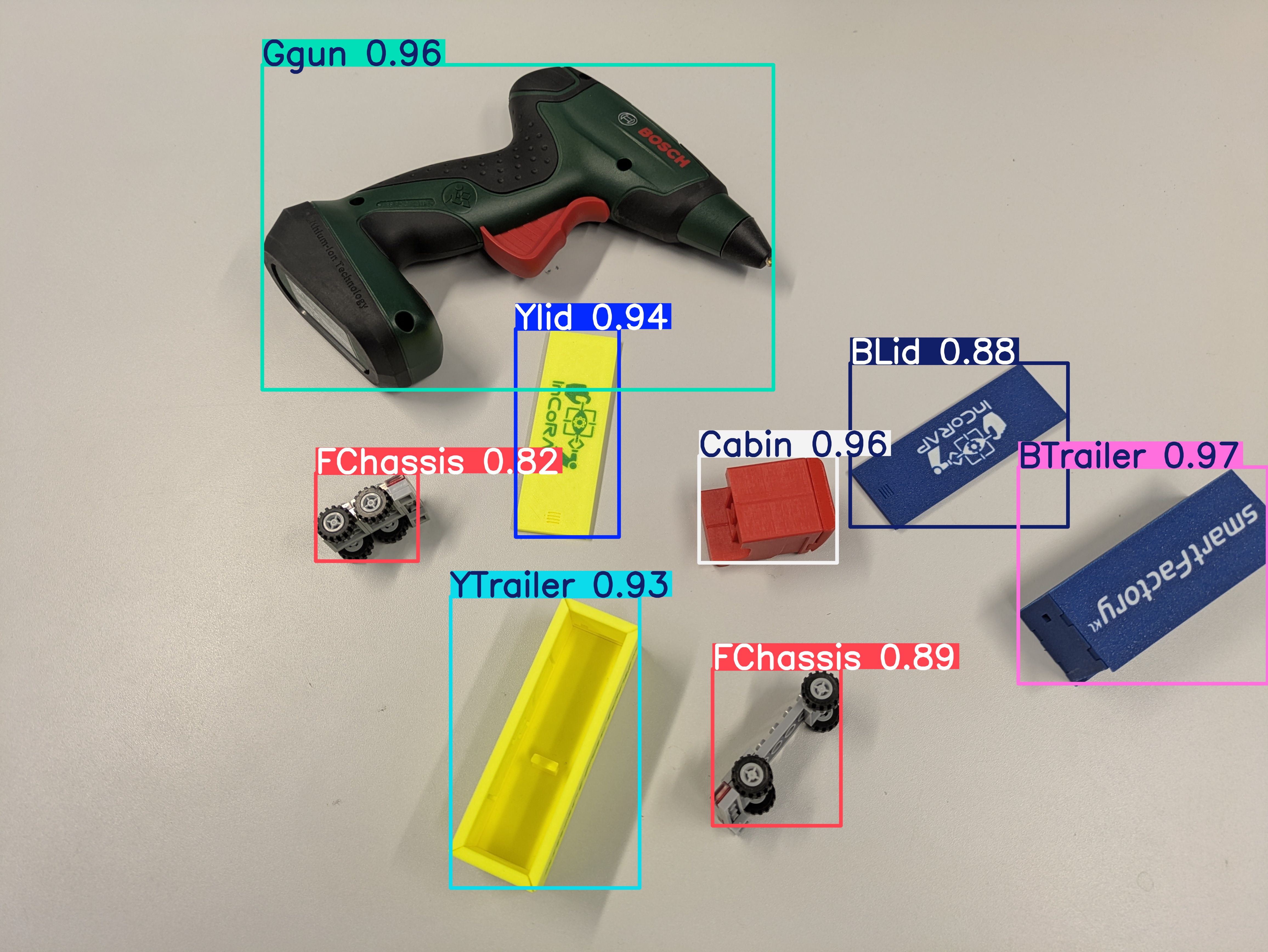}
        \caption{Experiment 1 results}
        \label{fig:exp1-cam2-M}
    \end{subfigure}
    \hfill
    \begin{subfigure}[t]{0.31\textwidth}
        \centering
        \includegraphics[width=\textwidth]{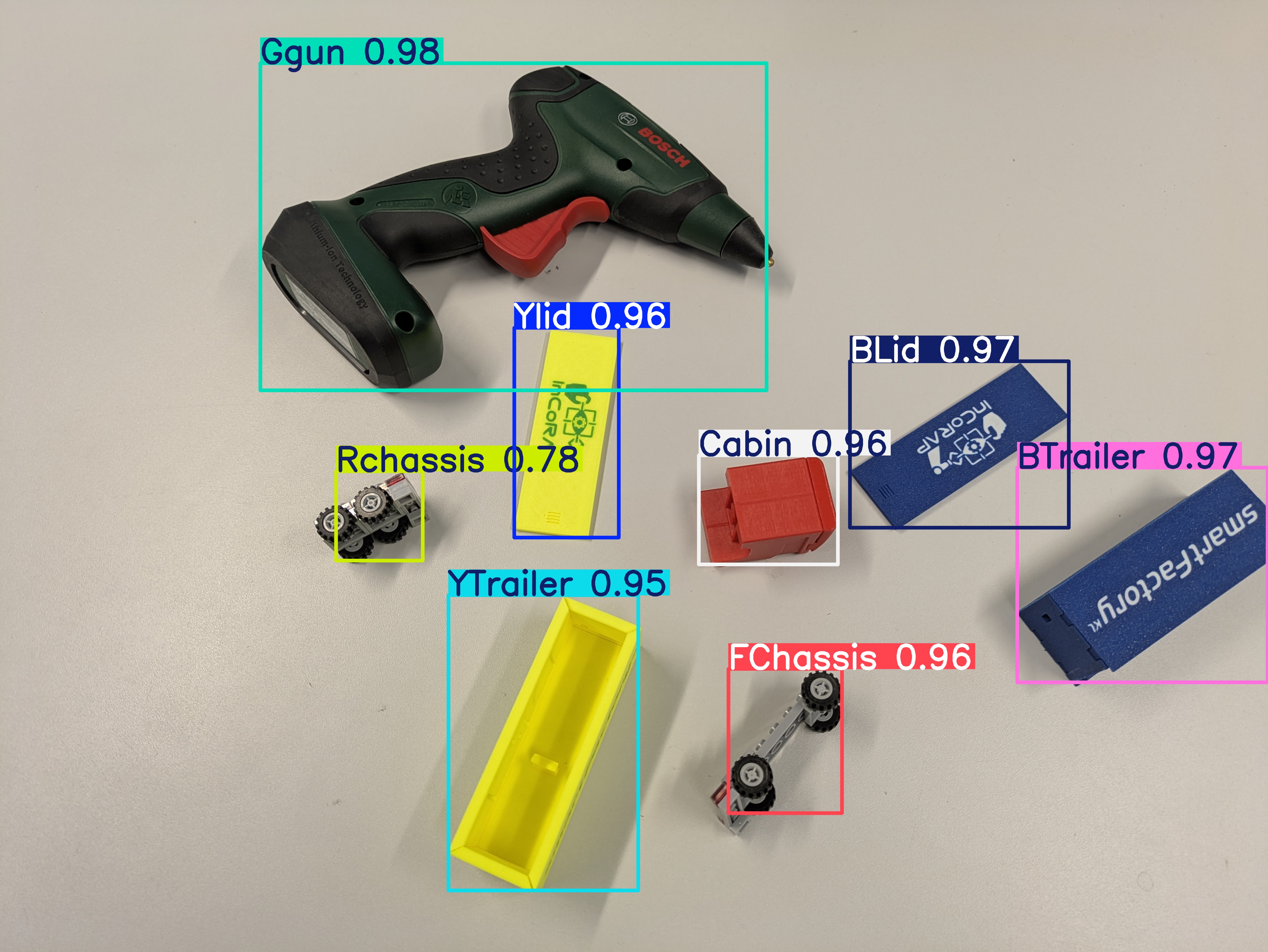}
        \caption{Experiment 2 results}
        \label{fig:exp2-cam2-M}
    \end{subfigure}
    
    \caption{Comparison of OD results for experiments 0, 1, and 2 under medium (M) light conditions using Camera 2.}
    \label{fig:OD-comparison-M}
\end{figure}

Figure~\ref{fig:camera_comparison} presents two controlled pairwise comparisons that isolate distinct variables within the SDG pipeline. Subfigure~\ref{fig:both_camA} compares Experiment~0 and Experiment~1 exclusively through Camera~2, isolating the effect of light transport type while keeping both background and camera geometry constant. Subfigure~\ref{fig:both_camB} compares Experiment~1 and Experiment~2 exclusively through Camera~1, isolating the effect of multi-bounce complexity while keeping both background and camera geometry constant. These two controlled comparisons provide the evidentiary foundation for interpreting the compound results of Experiment~2 Camera~2, which varies both background and lighting simultaneously and therefore cannot, in isolation, attribute its gains to either factor alone.

In Subfigure~\ref{fig:both_camA}, the sole difference between Experiment~0 and Experiment~1 through Camera~2 is the introduction of a white ground plane in Experiment~1, which adds the first indirect bounce — ceiling light reflects off the floor and illuminates the underside of the components before reaching the off-axis camera. At Low intensity, this single indirect bounce produces a gain of $+5.1$~pp ($55.8\%$ to $60.9\%$), confirming that even a minimal reflective surface meaningfully enriches the training images when direct illumination alone is insufficient to generate adequate feature gradients. At Medium intensity, the improvement narrows to $+1.7$~pp, as the stronger direct illumination already provides reasonable feature contrast and the marginal contribution of the ground-plane bounce diminishes. At High intensity, a slight reversal of $-1.0$~pp occurs: the high-energy ceiling light reflected off the ground plane introduces a secondary upward illumination component that partially disrupts surface shading, slightly degrading image quality relative to the unobstructed direct-light baseline. Critically, this comparison is free of any background; both conditions share an empty dark background as seen by Camera~2, and the entire $+5.1$~pp gain at Low intensity is attributable exclusively to the introduction of one indirect light bounce.

In Subfigure~\ref{fig:both_camB}, the sole difference between Experiment~1 and Experiment~2 through Camera~1 is the introduction of the full SmartFactoryKL virtual environment in Experiment~2, which enables multi-bounce indirect illumination from factory walls, shelving, and equipment. Both conditions share the same white ground-plane background as seen from Camera~1's top-down perspective. At Low intensity, Experiment~1 and Experiment~2 achieve identical mAP scores of $61.4\%$, a result that holds with equal precision for Camera~2 simultaneously ($60.9\%$), establishing a clear minimum illumination threshold as was mentioned, and the identical values across both cameras confirm this is a lighting energy constraint rather than a geometric artifact. At Medium intensity, a marginal gain of $+0.7$~pp is observed, indicating that the factory environment begins to contribute weakly at moderate energy levels but cannot overcome Camera~1's geometric susceptibility to the specular peak. At High intensity, a substantial drop of $-5.3$~pp ($55.3\%$ to $50.0\%$) demonstrates that multi-bounce PBS complexity without an off-axis camera geometry is actively detrimental: high-intensity multi-surface reflections compound the coaxial specular saturation effect already present in Camera~1, degrading performance well below Experiment~1. The contrast between Subfigure~\ref{fig:both_camA} ($-1.0$~pp at High, single reflective surface) and Subfigure~\ref{fig:both_camB} ($-5.3$~pp at High, many reflective surfaces) confirms that the severity of High-intensity degradation scales directly with the number of indirect reflective surfaces in the scene and is further amplified when the camera geometry is co-axial with the dominant light source.

The compound gains of Experiment~2, Camera~2, $+8.7$~pp at Medium and $+15.6$~pp at High over Experiment~1, Camera~2, reflect the interaction of two simultaneous changes: the multi-bounce PBS lighting enriching the diffuse radiance on object surfaces, and the complex SmartFactoryKL background adding scene-level training diversity that closes the domain gap between synthetic and real images. Because these two factors change simultaneously in this comparison, the individual contribution of each cannot be cleanly separated. However, the controlled comparisons in Sub-figures~\ref{fig:both_camA} and~\ref{fig:both_camB} establish that lighting type alone contributes up to $+5.1$~pp and that PBS complexity alone with Camera~1 contributes negligibly or negatively. The substantial additional gains in Experiment~2 Camera~2 therefore arise from the \emph{interaction} of two factors — the indirect multi-bounce PBS that requires Camera~2's off-axis geometry to be exploited, combined with background diversity that Camera~2's broader field of view captures more effectively. At Low intensity, both cameras simultaneously confirm no additional gain from the full factory environment, providing a boundary condition that applies equally to the lighting and background contributions: neither can produce informative training variation without sufficient illumination.

The per-class results in Table~\ref{tab:results_illustration_improved} and Figure~\ref{fig:avrage_chart_classwise_cam2} provide evidence for why experimental performance varies so dramatically across object classes. The \textbf{GlueGun}, the largest-size component in the dataset with the richest multi-material surface geometry, achieves the highest absolute performance of any class, peaking at 86.8\% mAP under Experiment~2 Camera~2 on average of light profiles. Its large physical footprint maximizes the number of texture-bearing pixels in each training image, and its complex multi-material BRDF produces a rich diffuse-lobe response that indirect multi-bounce PBS illuminates comprehensively.
The \textbf{RChassis} and \textbf{FChassis} components, carrying harsh and multi-layered surface textures, exhibit the largest experiment-to-experiment variation in the study. In Experiment~0, RChassis falls to 7.1\% (Camera~2 Low intensity) and 9.1\% mAP (Camera~1 High intensity), the two lowest per-class results, representing opposite failure modes of illumination. Insufficient radiance at Low intensity suppresses surface gradients below the detection threshold, while specular peak saturation at High intensity with coaxial geometry clips pixel values and equally eliminates the texture information the backbone requires. The same component recovers to 56.8\% in Experiment~2, Camera~2, High intensity, a gain exceeding 49~percentage points. Given that RChassis and FChassis share structurally similar chassis geometry, part of the residual error under poor illumination conditions may stem from inter-class confusion between these two visually related components rather than complete failure to detect an object.

The \textbf{Cabin} class presents the study's primary exception: as illustrated in Figure~\ref{fig:avrage_chart_classwise_cam2}, it is the only component where Experiment~0 and Experiment~1 outperform Experiment~2 on average. Critically, this is not a general sensitivity to low illumination — Cabin performs well under Low intensity in both Experiment~0 (65.5\%) and Experiment~1 (71.3\%), with the empty background. The collapse occurs specifically when Low intensity is combined with the complex factory background of Experiment~2, where detection falls to 11.3\% through Camera~2, while Medium and High intensity in the identical camera and background configuration recover to 72.5\% and 78.5\% respectively. This contrast demonstrates that the failure is not attributable to illumination intensity alone, nor to background complexity alone, but to their \textit{interaction}: weak indirect radiance is insufficient when it must also resolve a cluttered, texture-rich background simultaneously with the object. In simpler backgrounds, even Low intensity provides enough contrast for the model to separate the object from the scene; in the complex SmartFactoryKL environment, the same illumination budget is no longer sufficient to distinguish the Cabin's surface from the surrounding clutter. The Cabin's small size and multi-faceted surface, whose silhouette changes substantially with viewing angle, make it disproportionately vulnerable to this interaction compared to larger, more geometrically stable components. This finding refines the minimum-illumination-threshold mechanism observed for RChassis and FChassis: the threshold is not a fixed property of lighting intensity alone, but is modulated by background complexity, and is most acutely exposed by small, multi-faceted objects with high viewpoint variance.

Taken together, these findings demonstrate that optimizing synthetic training datasets for industrial object detection requires the co-design of three interdependent variables: camera geometry must be positioned off-axis relative to dominant light sources to access the diffuse BRDF lobe; illumination must use indirect multi-bounce PBS to enrich that lobe with object-surface texture information; and scene backgrounds must introduce domain-relevant environmental complexity at sufficient light intensity to produce informative rather than degenerative training variation. The illumination-background interaction threshold, the per-class texture gradient, and the interaction between camera geometry and lighting profile are all grounded in the rendering equation and collectively explain the performance distribution observed across the 18 experimental configurations.

\clearpage

\section{Conclusion}

In this study, we present an evaluation of the YOLOv12 object detection model to investigate its behavior when trained under varying synthetic data conditions. Specifically, we focus on the impact of different lighting configurations and their influence on model performance, aiming to identify conditions that yield the highest performance. Furthermore, the experiments were designed to demonstrate how increasing the complexity of lighting conditions, including reflections from components within the virtual environment, can enhance the detection capability of the object detection model. Moreover, the findings indicate that incorporating complex backgrounds further improves detection performance, with the effect becoming even more pronounced when sufficient illumination is jointly applied to both the target component and the background; Below this illumination threshold, the same background complexity instead degrades model convergence.

To validate these assumptions and implement our experiments, we developed SmartSDG, a synthetic data generation pipeline built on NVIDIA Isaac Sim, tailored for generating SmartFactoryKL-related scenarios for OD tasks. In addition, we created ILLUM\_INTRUCK, the synthetic dataset for an assembly line setting, and evaluated the trained models against a corresponding real-world dataset.
Despite the real test set being captured under approximately medium illumination, the highest mAP is achieved by the models trained under High intensity with indirect multi-bounce PBS, a configuration that does not match the test illumination level. This demonstrates that the benefit of PBS indirect lighting is not contingent on intensity matching between synthetic training and real evaluation conditions. Physically-based multi-bounce illumination creates feature representations that are robust to intensity variation, whereas direct lighting fails to optimally generalize even when the training intensity is nominally similar to the real test conditions.
Our findings demonstrate that combining complex, natural lighting conditions with domain-relevant backgrounds enhances the robustness of the model while simultaneously reducing false positives in detection.

Ultimately, utilizing the appropriate lighting configurations and complex environments in synthetic datasets proves to be an effective strategy for rendering target components more realistic, thereby enabling the object detection model to generalize more effectively to the target domain.
By using PBS (via NVIDIA Isaac Sim), the synthetic truck parts don't just look like "colored shapes." They react to the SmartFactoryKL lighting intensities like real materials and components. This physical accuracy is what allows the object detection model to learn features that actually exist in the real world, effectively closing the domain gap.

\section{Future Work}

The present study identifies surface texture density as a significant factor in determining how much an object class benefits from indirect PBS illumination — with high-texture components such as RChassis gaining over $49$~pp across the experimental space while smooth-surfaced components show smaller improvements. However, the systematic relationship between specific material properties, object geometry, and optimal lighting configuration was not deeply investigated. Future work should conduct a controlled per-material study, varying surface reflectance models, texture frequency, and object scale independently to establish principled guidelines for selecting lighting configurations based on component properties rather than empirical observation alone. This would extend the rendering equation framework developed here toward a predictive model for synthetic data quality.

A second important direction concerns out-of-distribution lighting evaluation. The real test set in this study was captured under approximately medium illumination conditions, meaning all 18 trained models were evaluated against a single real-world lighting scenario. Future work should collect or synthesize real test sets spanning a broader range of illumination levels, including low-light and high-intensity conditions, to evaluate how models trained under each experimental configuration respond when the real deployment environment deviates significantly from the training distribution. This would directly test whether indirect PBS training provides greater robustness to real-world illumination variation than direct lighting, a hypothesis supported by the current results but not directly measured.

The minimum illumination threshold identified for high-texture objects in complex backgrounds represents a specific design constraint that warrants further investigation. Future work should systematically characterise this threshold across a wider range of object textures, background complexities, and indirect lighting configurations to determine whether it follows a predictable relationship with surface spatial frequency and scene energy budget, which would enable automated lighting parameter selection in SDG pipelines.

Finally, the findings of this study are currently validated on a single industrial testbed, a single detector architecture, and a fixed camera geometry set. Extending the controlled experimental framework to dynamic environments, a wider range of object classes, outdoor scenarios, and additional object detection architectures would establish the generalisability of the co-design principle across broader synthetic data generation tasks.

\clearpage

\begin{table}[ht]
\centering
\caption{Comparison of SmartSDG configurations for single-object and multi-object dataset generation.}
\label{tab:smartsdg}
\begin{threeparttable}
\begin{tabular}{p{3.5cm}p{5.5cm}|p{5.5cm}}
\hline
\textbf{Category} & \textbf{Single-object mode} & \textbf{Multi-object mode} \\ \hline

\textbf{Environment setup} & 
Supports SmartFactory environment, flat background, or empty scene. \newline 
Headless execution mode for reproducible, large-scale generation. & 
Same setup; optimized for faster rendering and larger batch generation. \\[6pt]
\hline
\textbf{Cameras \& Rendering} & 
Two cameras: ceiling (top-down) and orbital side view. \newline 
Focal length: $48.0$, \newline Focus distance: $400$.  \newline 
Resolution: $1280 \times 800$.  \newline 
Path-tracing renderer with denoiser and anti-aliasing.  \newline 
Samples per pixel: $256$.  \newline 
Max bounces: $32$. & 
Two cameras: same configuration.  \newline 
\newline
\newline
\newline
Resolution: $1138 \times 640$.  \newline 
Path-tracing renderer with denoiser and anti-aliasing.  \newline 
Samples per pixel: $32$.  \newline 
Max bounces: $16$. \\[6pt]
\hline
\textbf{Lighting \tnote{*}} & 
 
\textbf{Intensity}: (Lux) 
\newline \textit{L:} $\mathcal{N}(5000,500)$ \newline \textit{M:} $\mathcal{N}(20000, 5000)$ \newline \textit{H:} $\mathcal{N}(80000,20000)$  
\newline
\textbf{Temperature}: (Kelvin) 
\newline $\mathcal{N}(8000,500)$.  
\newline
\textbf{16 fixed positions for overhead rectangular illumination.} & 
Same lighting setup as single-object mode. \\[6pt]
\hline
\textbf{Objects} & 
Single target object per image.  \newline
Components: \textit{Cabin}, \textit{GlueGun}, \textit{Chassis}, \textit{Trailer}, \textit{Lid}, etc.  \newline
Randomized translation, rotation ($\pm 180^\circ$), scaling. & 
Multiple objects per image with the same components and randomization. \\[6pt]
\hline
\textbf{Scene variation} & 
Experiment~0:~Object-only. \newline 
Experiment~1:~Ground plane with reflections.  \newline
Experiment~2:~Full SmartFactoryKL environment with clutter/complex background. & 
The same experimental variations were applied in multi-object scenes. \\[6pt]
\hline
\textbf{Output data} & 
RGB images, \newline YOLO-format bounding boxes.  \newline
Automatic filtering of samples with occlusion $>50\%$.  \newline
Optional semantic segmentation/metadata. & 
Same output pipeline and filtering mechanism. \\[6pt]
\hline
\textbf{Reproducibility} & 
Deterministic via random seed.  \newline
Organized output directory by class and lighting mode. & 
Same reproducibility pipeline. \\ \hline

\end{tabular}
\begin{tablenotes}
\footnotesize
\item[*] {These hyperparameters are empirically calibrated based on conventional SmartFactoryKL synthetic data generation (SDG) workflows. While the Medium (M) configuration establishes a nominal baseline replicating standard factory lighting, the Low (L) and High (H) profiles establish out-of-distribution parameters to evaluate robust model generalization. Notably, the standard deviations scale proportionally with the mean profiles to accurately simulate heteroscedastic real-world environmental fluctuations across different lighting extremes.}
\end{tablenotes}
\end{threeparttable}
\end{table}

\clearpage

\begin{table*}[htbp]
\centering
\setlength{\tabcolsep}{5pt}
\renewcommand{\arraystretch}{1.15}
\footnotesize
\caption{%
  YOLOv12 mAP[.50:.95] (in percentage) per class across all 18 experimental conditions.
  Columns are grouped by \textbf{Experiment} (0--2) and \textbf{Camera} (C1: coaxial;
  C2: off-axis), each split into three light intensities:
  \textbf{L}~(Low), \textbf{M}~(Medium), \textbf{H}~(High).
  \colorbox{bestcell}{\textbf{Bold green}} = highest score per class row;
  \colorbox{worstcell}{red} = lowest score per class row.
}
\label{tab:results_illustration_improved}
\begin{tabular}{%
  l
  ccc @{\hskip 6pt} ccc
  @{\hskip 10pt}
  ccc @{\hskip 6pt} ccc
  @{\hskip 10pt}
  ccc @{\hskip 6pt} ccc
}
\toprule
& \multicolumn{6}{c}{\cellcolor{hdr0}\textbf{Experiment 0}}
& \multicolumn{6}{c}{\cellcolor{hdr1}\textbf{Experiment 1}}
& \multicolumn{6}{c}{\cellcolor{hdr2}\textbf{Experiment 2}} \\
\cmidrule(lr){2-7}\cmidrule(lr){8-13}\cmidrule(l){14-19}
& \multicolumn{3}{c}{\textit{Camera 1}}
& \multicolumn{3}{c}{\textit{Camera 2}}
& \multicolumn{3}{c}{\textit{Camera 1}}
& \multicolumn{3}{c}{\textit{Camera 2}}
& \multicolumn{3}{c}{\textit{Camera 1}}
& \multicolumn{3}{c}{\textit{Camera 2}} \\
\cmidrule(lr){2-4}  \cmidrule(lr){5-7}
\cmidrule(lr){8-10} \cmidrule(lr){11-13}
\cmidrule(lr){14-16}\cmidrule(l){17-19}
\textbf{Class}
& L & M & H  & L & M & H
& L & M & H  & L & M & H
& L & M & H  & L & M & H \\
\midrule
\rowcolor{allrow}
\textbf{All}
  & 43.1 & 45.5 & \W{35.7}   & 55.8 & 62.1 & 61.2
  & 61.4 & 64.9 & 55.3       & 60.9 & 63.8 & 60.2
  & 61.4 & 65.6 & 50.0       & 60.9 & 72.5 & \B{75.8} \\
\midrule
BLid
  & \W{16.7} & 49.3 & 27.7   & 66.5 & 66.8 & 74.0
  & 68.2 & 68.5 & 54.7       & 68.6 & 69.6 & 71.7
  & 68.2 & 63.6 & 52.0       & 68.6 & 75.3 & \B{79.1} \\
BTrailer
  & 46.9 & 51.0 & \W{19.7}   & 60.9 & 73.2 & 67.9
  & 73.3 & 71.6 & 62.3       & 76.5 & 71.7 & 65.5
  & 73.3 & 70.0 & 42.5       & 76.5 & \B{78.6} & 77.5 \\
Cabin
  & 40.3 & 70.7 & 68.7       & 65.5 & 73.9 & 75.0
  & 70.2 & 74.7 & 72.8       & 71.3 & 71.1 & 71.5
  & 20.3 & 74.6 & 69.4       & \W{11.3} & 72.5 & \B{78.5} \\
FChassis
  & \W{16.9} & 31.3 & 26.6   & 44.5 & 50.3 & 45.3
  & 51.2 & 53.2 & 48.0       & 52.8 & 50.8 & 50.2
  & 55.2 & 55.9 & 41.7       & 57.7 & 63.7 & \B{65.8} \\
GlueGun
  & 72.2 & 41.2 & \W{26.0}   & 70.5 & 79.1 & 65.5
  & 79.2 & 63.0 & 60.9       & 80.0 & 80.2 & 68.1
  & 79.2 & 69.8 & 60.2       & 87.4 & \B{88.3} & 84.6 \\
RChassis
  & 22.3 & 24.5 & 9.1        & \W{7.1} & 18.1 & 27.7
  & 48.1 & 39.0 & 14.1       & 41.6 & 26.5 & 18.5
  & 48.2 & 49.3 & 12.8       & 41.6 & 39.6 & \B{56.8} \\
YTrailer
  & 69.8 & \W{43.3} & 52.3   & 60.9 & 71.4 & 69.5
  & 76.4 & 77.1 & 62.8       & 74.9 & 72.1 & 69.1
  & 76.4 & 72.1 & 55.6       & 74.9 & 84.5 & \B{85.1} \\
YLid
  & 59.9 & \W{52.7} & 55.6   & 59.5 & 63.8 & 64.4
  & 70.3 & 71.9 & 67.0       & 68.7 & 68.7 & 67.0
  & 70.3 & 69.2 & 65.4       & 68.7 & 77.5 & \B{78.9} \\
\bottomrule
\end{tabular}
\end{table*}
\clearpage

\usetikzlibrary{shapes.geometric, arrows.meta, positioning, calc, fit, backgrounds}

\bibliographystyle{unsrtnat}
\bibliography{sample}

\end{document}